\documentclass[10pt,twocolumn,letterpaper]{article}

\usepackage{iccv}              

\usepackage[accsupp]{axessibility}
\usepackage{multirow}
\usepackage{makecell}
\usepackage{arydshln}
\usepackage{graphicx}
\usepackage{fontawesome}
\usepackage{bbding}
\usepackage{pifont}
\usepackage{cuted}
\usepackage{amsmath}
\usepackage{algorithm}
\usepackage{algpseudocode}
\usepackage{amsfonts}
\usepackage{amssymb}
\usepackage{amssymb}

\definecolor{iccvblue}{rgb}{0.21,0.49,0.74}
\usepackage[pagebackref,breaklinks,colorlinks,allcolors=iccvblue]{hyperref}

\def\paperID{3061} 
\def\confName{ICCV}
\def\confYear{2025}

\title{StrandHead: Text to Hair-Disentangled 3D Head Avatars \\ Using Human-Centric Priors}
\vspace{-2cm}
\author{
    Xiaokun Sun, 
    Zeyu Cai, 
    Ying Tai, 
    Jian Yang,
    Zhenyu Zhang\thanks{Corresponding author}\\
    Nanjing University \quad $^{*}$Corresponding Author\\
    {\tt\small xiaokun\_sun@smail.nju.edu.cn, caizeyu010612@gmail.com,} \\
    {\tt\small \{yingtai, csjyang\}@nju.edu.cn, zhangjesse@foxmail.com}
    }

\begin{document}
\twocolumn[{
\renewcommand\twocolumn[1][]{#1}
\maketitle
\vspace{-14mm}
\begin{center}
    \captionsetup{type=figure}
    \includegraphics[width=0.83\textwidth]{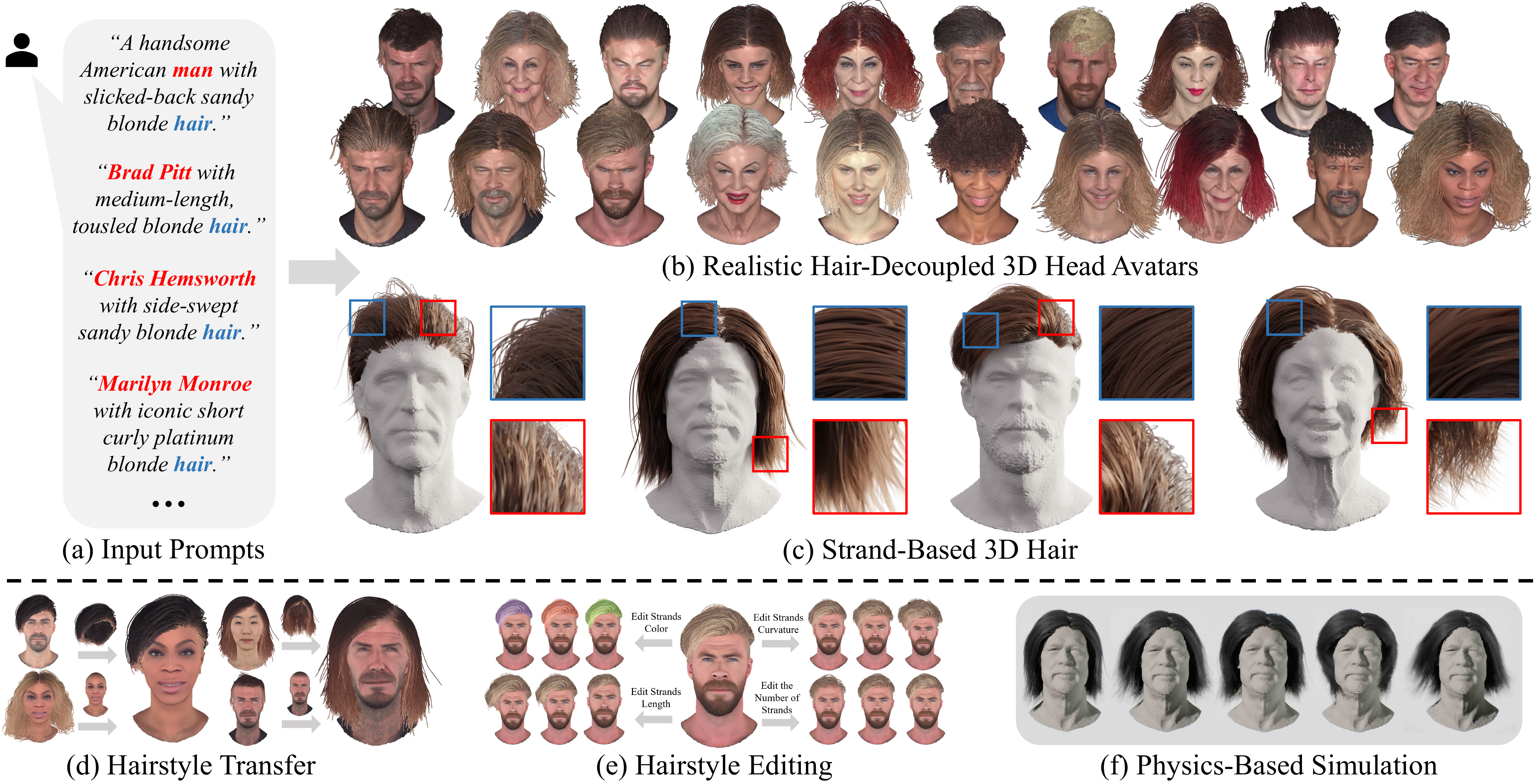}
    \vspace{-3mm}
    \caption{Given \textbf{(a)} input prompts, \textbf{StrandHead} generates \textbf{(b)} realistic 3D head avatars featuring strand-level attributes and \textbf{(c)} 3D hair strands by utilizing human-specific 2D generative priors and 3D hair strand geometric priors.
    By precisely capturing the internal geometry of hair strands, our approach enables seamless and flexible \textbf{(d)} hairstyle transfer and \textbf{(e)} editing, as well as \textbf{(f)} physics-based simulation.}
    \label{fig:1.1}
\end{center}
}]

\begin{abstract}

While haircut indicates distinct personality, existing avatar generation methods fail to model practical hair due to the data limitation or entangled representation. We propose StrandHead, a novel text-driven method capable of generating 3D hair strands and disentangled head avatars with strand-level attributes. Instead of using large-scale hair-text paired data for supervision, we demonstrate that realistic hair strands can be generated from prompts by distilling 2D generative models pre-trained on human mesh data. To this end, we propose a meshing approach guided by strand geometry to guarantee the gradient flow from the distillation objective to the neural strand representation. The optimization is then regularized by statistically significant haircut features, leading to stable updating of strands against unreasonable drifting. These employed 2D/3D human-centric priors contribute to text-aligned and realistic 3D strand generation. Extensive experiments show that StrandHead achieves the state-of-the-art performance on text to strand generation and disentangled 3D head avatar modeling. The generated 3D hair can be applied on avatars for strand-level editing, as well as implemented in the graphics engine for physical simulation or other applications.
Project page: \url{https://xiaokunsun.github.io/StrandHead.github.io/}.

\end{abstract}    
\vspace{-0.5cm}
\section{Introduction}
\label{sec:1}

Creating 3D head avatars is crucial for many applications including digital telepresence, gaming, movies, and AR/VR. Hairstyles reflecting personal characteristics greatly impact the fidelity and realism of digital humans. \textcolor{black}{Traditional methods rely heavily on manual effort, making them time-consuming and labor-intensive. Recent advances~\cite{coma,i3dmm,imface,ganhead} achieve automatic 3D head generation with the research paradigm based on supervised learning. However, these methods require costly 3D training data, which significantly limits their generalization capabilities and restricts potential applications.}

With the rapid development of \textcolor{black}{text-driven generation methods~\cite{latentdiffusion, dreamfusion}, creating 3D head avatars from given prompts without requiring 3D-text datasets becomes possible. Recent studies~\cite{clipface,headsculpt,headstudio,teca} unlock the potential of 2D diffusion models in modeling 3D head avatars by integrating them with 3D head priors~\cite{3dmm,flame,facescape}.
HumanNorm~\cite{humannorm} further improves domain-specific fidelity by fine-tuning diffusion models on high-quality 3D human meshes.
While focusing on facial geometry and texture modeling, these methods utilize holistic meshes or NeRF~\cite{nerf} to represent 3D haircuts, failing to capture the internal geometric structure of hair strands, i.e., 3D curves. This limitation not only significantly reduces the realism of the generated avatars but also makes them incompatible with strand-based applications and simulation systems~\cite{blender,maya,unrealengine}.}


\textcolor{black}{To accurately model 3D haircuts, recent efforts achieve strand-level hairstyle reconstruction~\cite{neural_strands,neural_haircut,hairstep} or generation~\cite{groomgen, perm} using VAEs or parametric models. Despite their impressive performance, these methods require constrained multi-view images or manual latent space searching, preventing them from freely creating 3D hair based on user-friendly text.
The most recent work, HAAR~\cite{haar} pioneers text-to-strand by training a text-conditioned hair map diffusion model. However, due to its reliance on large-scale and costly paired data including 9825 haircuts and descriptions, HAAR faces challenges in generating diverse hairstyles outside the training set.  Additionally, HAAR overlooks hair texture and geometry that adapts to specific head shapes, further limiting its practical applications.}


So can we use powerful human-specific 2D generative priors to create realistic and diverse strand-based hair from text?
\textcolor{black}{We address this meaningful and challenging problem by developing \textbf{StrandHead}, a novel framework that generates high-fidelity 3D head avatars with strand-accurate hair from prompts.
As illustrated in \cref{fig:1.1}, the generated head avatars feature diverse and realistic hair strands, enabling seamless and flexible strand-level transfer, editing, and physics-based simulation.
Instead of using large-scale hair-prompt paired data, we achieve text-aligned 3D hair strand generation by effectively utilizing human-specific 2D generative priors and 3D hair geometric priors.}

Specifically, we propose a novel differentiable prismatization algorithm inspired by the cylindrical structure of hair strands. This algorithm can efficiently and differentiably convert strands into watertight prismatic meshes, which enables smooth backpropagation of gradients from 2D diffusion models pre-trained on human mesh data to 3D hair strands using mesh-based differentiable renderers, thereby making it possible to use 2D generative priors for modeling 3D hairstyles.
Then we propose two effective losses inspired by the distribution patterns of 3D hair geometric features to further regularize the hair shape.
These rich 2D/3D human-centric priors work together to achieve text-aligned and realistic 3D hair strand generation. 
Extensive experiments demonstrate that StrandHead outperforms the state-of-the-art (SOTA) methods in both head and hair generation tasks, and supports flexible haircut transfer and editing, as well as physical-based rendering and simulation.

Our main contributions are summarized as follows:
\begin{itemize}
    \item We propose StrandHead, a novel framework for generating realistic 3D head avatars with strand-level attributes. 
    To the best of our knowledge, StrandHead is the first work to generate 3D hair strands by distilling human-specific 2D diffusion models.
    \item We propose a differentiable prismatization algorithm that converts hair strands into watertight prismatic meshes. This ensures stable gradient flow from 2D generative priors to 3D hair strand representation, thereby achieving reliable end-to-end strand-based hair optimization.
    \item Inspired by statistical 3D hair geometric features, we introduce two simple but solid regularization losses to supervise both local and global hair shapes, enabling reasonable and realistic hairstyle generation.
\end{itemize}

\section{Related Work}
\label{sec:2}

\noindent
\textbf{Text-to-3D General Object Generation.}
Inspired by the success of text-to-image (T2I) generation~\cite{t2i_1,t2i_2, t2i_3,t2i_4,latentdiffusion,controlnet}, many studies have explored using pre-trained vision-language models~\cite{clip,latentdiffusion} to achieve text-guided 3D content generation without large-scale 3D-text paired data.
Early methods~\cite{clip-mesh,text2mesh,clip-forge,dream_fields,clip-nerf} employ the CLIP model~\cite{clip} to supervise the alignment of 3D representations with prompts.
DreamFusion~\cite{dreamfusion} introduces the Score Distillation Sampling (SDS) loss, significantly enhancing the fidelity of generated 3D content by leveraging more powerful pre-trained diffusion models~\cite{latentdiffusion}. 
Subsequent works further advance text-driven 3D generation by improving 3D representations~\cite{fantasia3d,gsgen,gaussiandreamer}, optimization strategies~\cite{magic3d,dreamgaussian,gaussiandreamerpro}, SDS loss~\cite{prolificdreamer,luciddreamer}, and diffusion models~\cite{sweetdreamer,richdreamer,mvdream}. 
Despite these advancements, current approaches focused on general content generation do not fully leverage the extensive prior knowledge of human heads and hair, limiting their ability to generate realistic, high-quality 3D head avatars with strand-based hair.

\noindent
\textbf{Text-to-3D Head Avatar Generation.}
CLIPFace~\cite{clipface}, T2P~\cite{T2P} and Describe3D~\cite{describe3D} pioneer zero-shot text-driven 3D head generation by combining CLIP~\cite{clip} with 3D parametric head models~\cite{3dmm,flame,headscape,facescape}.
With the introduction of SDS loss~\cite{dreamfusion}, DreamFace~\cite{dreamface} and FaceG2E~\cite{faceg2e} leverage pre-trained T2I diffusion models to greatly improve generation quality within specific domains.
HeadEvolver~\cite{headevolver} enhances the expressiveness of head mesh deformations by introducing vector fields, while HeadSculpt~\cite{headsculpt}, and HeadArtist~\cite{headartist} employ DMTet~\cite{dmtet} instead of traditional meshes to capture geometric details. HeadStudio~\cite{headstudio} incorporates 3DGS~\cite{3dgs} to produce realistic and animatable 3D heads.
HumanNorm~\cite{humannorm}, further enhances human-specific fidelity by fine-tuning 2D diffusion models on high-quality human mesh data.
Despite these advancements, these approaches treat the head and hair as a holistic model, limiting support for downstream applications such as hairstyle transfer and editing.
To enable flexible head-hair-disentangled generation, TECA~\cite{teca} independently represents the head with mesh and the hair with NeRF~\cite{nerf}.
However, these methods focus only on the realistic external appearance of hair without modeling its internal geometric structure, restricting their application for strand-level editing and physics-based simulation.

\begin{table}[tp]
  \centering
  \resizebox{0.8\linewidth}{!}{
  \begin{tabular}{cccccc}
    \toprule
    Task & Method & \makecell[c]{Head-Hair-\\Decoupled} & \makecell[c]{Strand-Based\\Hair} & \makecell[c]{Geometry \&\\Texture} & \makecell[c]{No Large-Scale \\ 3D-Text Paired Data} \\
    \midrule
    \makecell[c]{Text-to-Head} & \makecell[c]{\cite{headsculpt,headartist,headstudio,humannorm}\\\cite{teca}} & \makecell[c]{\ding{55}\\\ding{51}} & \makecell[c]{\ding{55}\\\ding{55}} & \makecell[c]{\ding{51}\\\ding{51}} & \makecell[c]{\ding{51}\\\ding{51}}\\
    \hdashline
    \makecell[c]{Text-to-Hair} & \makecell[c]{\cite{mvdream,gaussiandreamer,luciddreamer,richdreamer}\\\cite{haar}} & \makecell[c]{\ding{55}\\\ding{55}} & \makecell[c]{\ding{55}\\\ding{51}} & \makecell[c]{\ding{51}\\\ding{55}} & \makecell[c]{\ding{51}\\\ding{55}}\\
    \hdashline
    \makecell[c]{Text-to-Head-Hair} & Ours & \ding{51} & \ding{51} & \ding{51} & \ding{51}\\
    \bottomrule
  \end{tabular}
  }
  \vspace{-0.3cm}
  \caption{Comparison with current related methods.}
  \vspace{-0.6cm}
  \label{tab:summary}
\end{table}

\noindent
\textbf{Strand-Based Hair Creation.}
Automatic strand-based hair modeling has garnered significant attention from both industry~\cite{blender,maya,unrealengine} and academia~\cite{hair_meshes,hair_refinement,real_hair,real_hair_inter}, as it enables downstream physics-based rendering and simulation
Such methods can be broadly categorized into two types: \textbf{(1)} reconstructing hair from input images or videos and \textbf{(2)} generating hair based on control conditions.
Works on the former one~\cite{sing_view_rec_hair,hairnet,hairvae,multi_view_rec_hair,deepmvshair,neural_strands,neuralhdhair,dynamic_hair,ct2hair,neural_haircut,hairstep,gaussianhair,gaussianhaircut,monohair,drhair,unihair,groomcap,hairmony} relies on classical 3D reconstruction frameworks~\cite{colmap,deepsdf,occ_net,nerf,3dgs} and incorporates hair-specific features~\cite{hair_capture} to realize accurate 3D strand reconstruction.
However, these methods depend on constrained multi-view images, making hair creation costly and challenging in less controlled settings.
Recently, 3D hair generation approaches~\cite{groomgen,perm} have emerged using GANs or parametric models, but they struggle with controllable generation.
The most relevant work to ours is HAAR~\cite{haar}, which pioneers text-to-strand by training a text-conditioned hair map diffusion model.
However, HAAR relies on a large-scale paired dataset including 9825 haircuts and prompts with limited diversity, restricting its ability to generate novel hairstyles beyond the training data.
Moreover, HAAR overlooks hair texture and geometry that adapts to specific head shapes, further limiting its application range.

Compared to previous methods, StrandHead generates 3D heads featuring strand-level hairstyles. Instead of requiring large-scale hair-text paired data, we achieve realistic and diverse 3D hair strand generation by using 2D/3D human-centric priors.
We summarize the main differences between our methods and related works in \cref{tab:summary}.
\begin{figure*}[tp]
    \centering
    \includegraphics[width=0.9\textwidth]{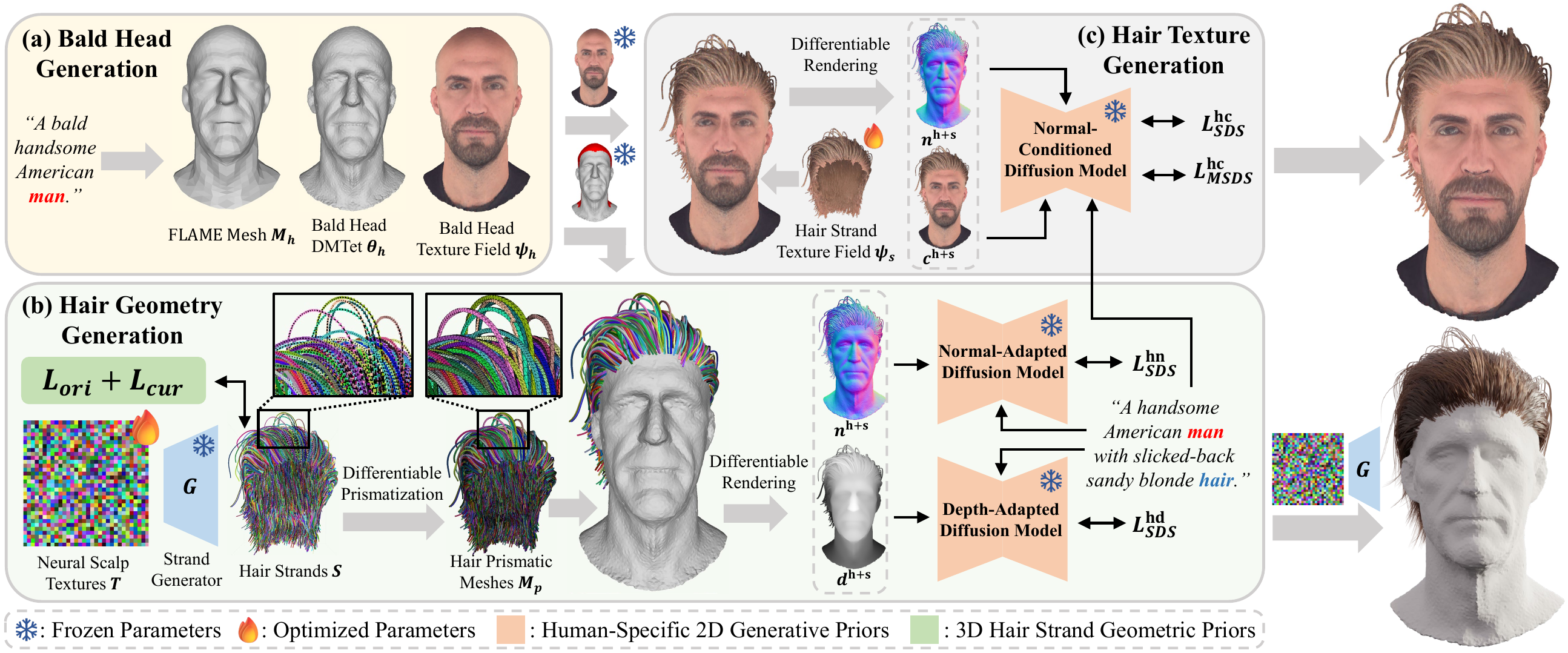} 
    \vspace{-0.3cm}
    \caption{StrandHead includes three stages: \textbf{(a)} We first create a FLAME-aligned 3D bald head using the improved HumanNorm~\cite{humannorm}.
    \textbf{(b)} Next, we introduce a differentiable prismatization algorithm to enable human-specific geometry-aware 2D diffusion models to supervise hair shape modeling. Additionally, two losses inspired by 3D hair geometric priors are applied to further regularize the hair geometry.
    \textbf{(c)} Finally, we use a human-specific normal-conditioned 2D diffusion model to generate lifelike hair textures.} 
    \label{fig:3.1}
    \vspace{-0.6cm}
\end{figure*}

\section{Method}
\label{sec:3}
\subsection{Preliminaries}
\label{sec:3.1}
\noindent
\textbf{FLAME}~\cite{flame} is a 3D parametric head model that represents the human head shape, pose, and expression using a compact set of parameters. 
Given a set of shape parameters $\beta$, pose parameters ${\theta}_\text{pose}$, and expression parameters $\psi_\text{exp}$, FLAME can create a 3D head mesh ${M}$. 

\noindent
\textbf{HumanNorm}~\cite{humannorm} is a text-driven, high-quality 3D head generation method. In specific, it models the geometry and texture of the human head using DMTet~\cite{dmtet} and a texture field.
The alignment between the text $y$ and the optimized representation $\theta$ is achieved using the following SDS loss~\cite{dreamfusion} with T2I diffusion models $\phi$:
\vspace{-0.3cm}
\begin{equation}
    \nabla_{\theta} \mathcal{L}_{SDS} = \mathbb{E}_{t, \epsilon} \left[ w(t) ({\epsilon}_{\phi}(x_t; y, t) - \epsilon ) \frac{\partial x}{\partial \theta} \right],
    \label{eq:sds}
    \vspace{-0.3cm}
\end{equation}
where $x = g(\theta)$ is the image rendered from $\theta$ by a differentiable renderer $g$, $t$ is the time step, $x_t = x + \epsilon$ is a noised version of $x$, ${\epsilon}_{\phi}(x_t; y, t)$ is the denoised image, and $w(t)$ is a weighting function.
HumanNorm employs human-specific diffusion models fine-tuned on high-quality human mesh data to replace the general diffusion model, enhancing the fidelity of the generated human heads.


\noindent
\textbf{Neural Scalp Textures (NST)}~\cite{neural_strands} is an efficient hair representation where each pixel stores a feature vector conveying the shape information of a single strand at the corresponding scalp location. 
Using a pre-trained hair strand generator~\cite{neural_strands} $G$, the low-dimensional 2D neural scalp texture $T$ can be decoded into high-dimensional 3D strand polylines $S = \{s^{i}\}_{i=1}^{N_{s}}$, where $N_{s}$ is the number of strands and $s^{i} = \{p^{i}_{j}\}_{j=1}^{N_{p}}$ consists of $N_{p}$ 3D points. 
This process is formulated as follows: $S = G(T)$.

\subsection{Overview}
\label{sec:3.2}
Given a text prompt, StrandHead aims to create a realistic 3D head avatar with strand-based hair without relying on large-scale 3D-text paired data.
\cref{fig:3.1} provides an overview of our three-stage pipeline. 
First, we generate a FLAME-aligned 3D bald head for accurate hair initialization using human-specific diffusion models (\cref{sec:3.3}).
In the subsequent stages, we model the reasonable geometry and realistic texture of 3D hair strands by leveraging rich 2D/3D human-centric priors (\cref{sec:3.4} and \cref{sec:3.5}).

\subsection{Bald Head Generation}
\label{sec:3.3}
\noindent
To achieve accurate hair initialization and optimize 3D hair strands in subsequent steps using human-specific 2D generative priors, we first need to obtain a reasonable and semantic-aligned 3D bald head from the text (\cref{fig:3.1}-(a)).
To achieve this goal, we enhance HumanNorm~\cite{humannorm} by incorporating an evolving FLAME~\cite{flame} model while optimizing the bald head DMTet $\theta_\text{h}$. This provides accurate semantic information and prevents unnatural geometry. Please refer to the Supp. Mat. for details on bald head generation.

\subsection{Hair Geometry Generation}
\label{sec:3.4}
\noindent
With a FLAME-aligned bald head, we can generate 3D hair in the scalp area. Specifically, we first initialize the haircut based on hairstyle descriptions, and then further sculpt the 3D hair shape utilizing 2D/3D human-centric priors.

\noindent
\textbf{Hair Initialization.}
To achieve reasonable and diverse hair initialization, we first utilize ChatGPT~\cite{chatgpt} to select the 20 most representative hairstyles from the USC-HairSalon Dataset~\cite{sing_view_rec_hair}.
We then optimize neural scalp textures to fit the selected hairstyle using the following loss function:
\vspace{-0.2cm}
\begin{equation}
    {\mathcal{L}}_\text{fit} = \sum^{N_{s}}_{i=1} \sum^{N_{p}}_{j=1} \Vert \hat{p}^{i}_{j} - p^{i}_{j} \Vert_{2} + \lambda_{\text{ori}} (1-\hat{o}^{i}_{j} \cdot o^{i}_{j}) + \lambda_{\text{cur}} \Vert \hat{c}^{i}_{j} - c^{i}_{j} \Vert_{1},
    \label{eq:init_hair}
\end{equation}
where $\hat{p}^{i}_{j}$ and ${p}^{i}_{j}$ are the position of the $j$-th point on the $i$-th polyline of the GT and generated hair, respectively, and $\hat{o}^{i}_{j}$, $o^{i}_{j}$, $\hat{c}^{i}_{j}$, and $c^{i}_{j}$ denote their orientation and curvature, respectively.
Here, orientation $o^{i}_{j} = (p^{i}_{j+1} - p^{i}_{j}) / \Vert p^{i}_{j+1} - p^{i}_{j} \Vert_{2}$ represents the direction of change in strand position, and curvature $c^{i}_{j} = \Vert o^{i}_{j} - o^{i}_{j-1} \Vert_{2}$ denote the rate of change in strand orientation.
Given a hair description, we select an optimal pre-trained NST as initialization and further optimize it in subsequent stages.

\begin{figure}[tp]
    \centering
    \includegraphics[width=0.45\textwidth]{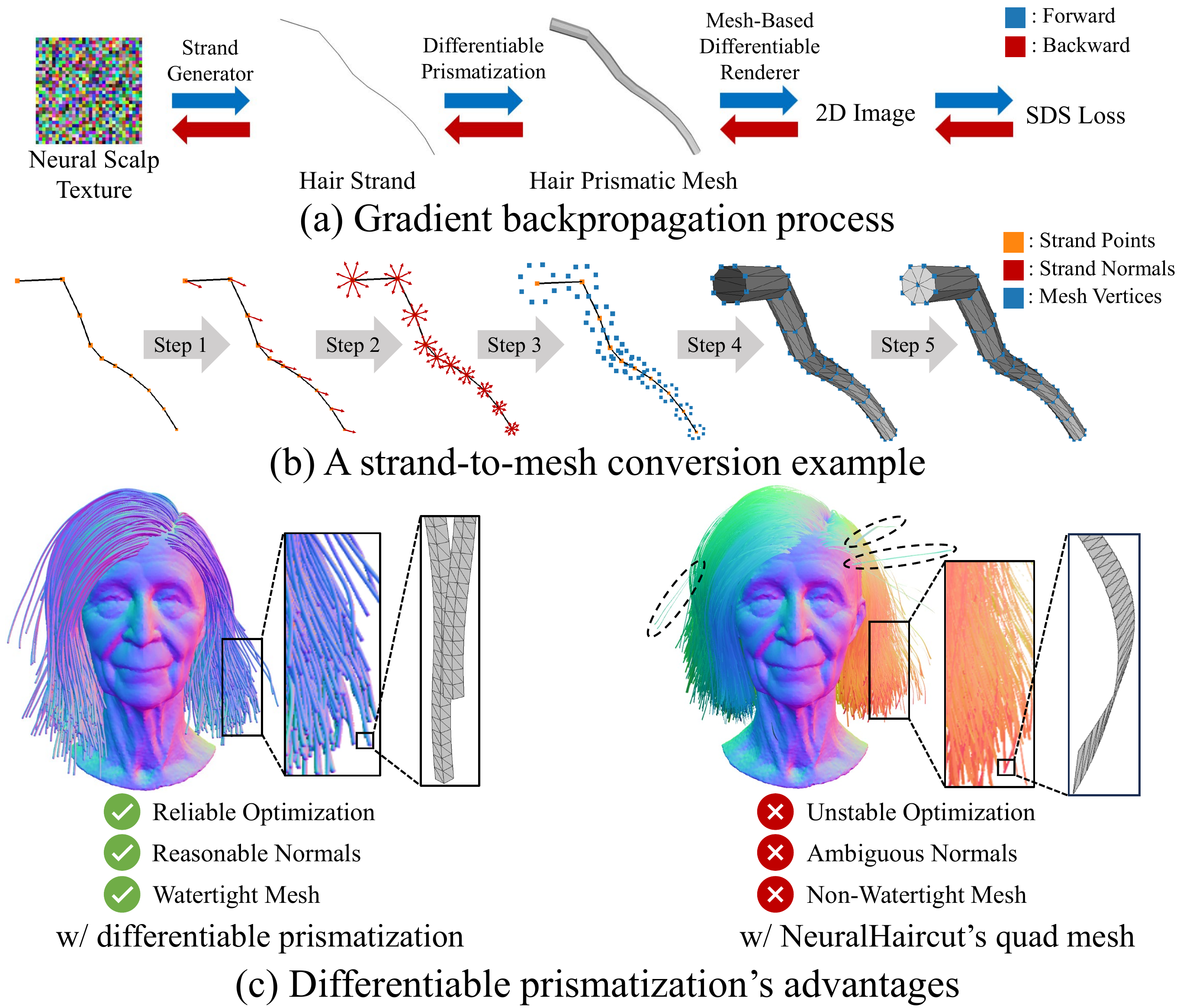} 
    \vspace{-0.3cm}
    \caption{The differentiable prismatization algorithm's \textbf{(a)} gradient backpropagation process, \textbf{(b)} a strand-to-mesh conversion example and \textbf{(c)} advantages over NeuralHaircut~\cite{neural_haircut}. Non-watertight quad meshes can easily produce ambiguous normal maps, which significantly reduce the stability of hair shape modeling (see the drifting hair highlighted by the oval dotted box in \textbf{(c)}).}
    \vspace{-0.6cm}
    \label{fig:3.2}
\end{figure}

\noindent
\textbf{Discussion on Strand-Based Differentiable Rendering.}
With reasonable hair initialization, the next step is to leverage human-specific 2D diffusion models to further sculpt the hair geometry.
However, due to the lack of a stable strand-based differentiable renderer, it is very challenging to optimize hair shape using 2D generative priors like other SDS-based methods.
Besides, we expect hair strands to produce mesh-style smooth geometry and texture maps to fully exploit the powerful generative capabilities of 2D diffusion models pre-trained on human mesh data.
A feasible alternative is to first differentiably convert the strands into quad meshes~\cite{hair_meshes}, similar to NeuralHaircut~\cite{neural_haircut}, and then leverage SDS-based mesh optimization frameworks to model hair strands.
However, this non-watertight stripe-like mesh \textcolor{black}{easily produces ambiguous normal maps or excessively thin sides (see the zoom image of \cref{fig:3.2}-(c))}, which \textcolor{black}{significantly} reduce the optimization stability (see the drifting hair highlighted by the oval dotted box in \cref{fig:3.2}-(c)).
Therefore, a differentiable strand-to-mesh conversion method with superior optimization properties is urgently needed.

\noindent
\textbf{Differentiable Prismatization (DP).}
To this end, we propose a novel differentiable prismatization algorithm inspired by the internal structure of 3D hair (i.e., hair fiber is a dielectric cylinder covered with tilted scales and with a pigmented interior~\cite{hair_rendering1,hair_rendering2}).
This algorithm can efficiently and differentiably convert strands into watertight prismatic meshes with arbitrary thickness and lateral edges to approximate the cylindrical structure of the hair.
This enables smooth backpropagation of gradients from the SDS loss to the 3D strand representation using mesh-based differentiable renderers~\cite{renderer}, thereby making it possible to fully exploit 2D generative priors distilled from high-fidelity human mesh data for modeling realistic hair strands. \cref{fig:3.2}-(a) visualizes the complete gradient backpropagation process.

In specific, given a hair strand $s$, our DP converts it into a watertight prismatic mesh with $K$ lateral edges and radius $R$ through the following five steps: \textbf{(1)} Compute the Initial Normal Vector. \textbf{(2)} Generate $K$ Rotated Normals. \textbf{(3)} Translate to Form Lateral Edges. \textbf{(4)} Construct Lateral Faces. \textbf{(5)} Construct Top and Bottom Faces. An example of this conversion is shown in \cref{fig:3.2}-(b). Unlike quad meshes~\cite{neural_haircut}, our DP effectively avoids ambiguous normals, enabling smooth gradient backpropagation and hair strand optimization, as shown in \cref{fig:3.2}-(c). Further algorithm details are provided in the Supp. Mat.

With the help of our proposed differentiable prismatization algorithm, we optimize the neural scalp texture $T$ by utilizing rich human-specific 2D generative priors through the following SDS losses (\cref{fig:3.1}-(b)):
\vspace{-0.3cm}
\begin{align}
    \nabla_{T} \mathcal{L}^{\text{hn}}_{SDS} &= \mathbb{E}_{t, \epsilon} \left[( {\epsilon}_{{\phi}_\text{hn}}(n^\text{h+s}_t; y_\text{h+s}, t) - \epsilon ) \frac{\partial n^\text{h+s}}{\partial T} \right],
    \label{eq:sds_hair_normal} \\
    \nabla_{T} \mathcal{L}^{\text{hd}}_{SDS} &= \mathbb{E}_{t, \epsilon} \left[( {\epsilon}_{{\phi}_\text{hd}}(d^\text{h+s}_t; y_\text{h+s}, t) - \epsilon ) \frac{\partial d^\text{h+s}}{\partial T} \right],
    \label{eq:sds_hair_depth}
    \vspace{-0.8cm}
\end{align}
where ${\phi}_\text{hn}$ and ${\phi}_\text{hd}$ are the human-specific normal-adapted and depth-adapted diffusion models from HumanNorm~\cite{humannorm}, $n^\text{h+s}$ and $d^\text{h+s}$ are the rendered normal and depth maps of the head DMTet $\theta_\text{h}$ with hair prismatic mesh $M_{p} = DP(G(T))$, respectively. Here, $DP(\cdot)$ denotes the differentiable prismatization operation function, $y_\text{h+s}$ is the full text description of the head with hair.

However, due to the excessive flexibility of hair strands, relying solely on the SDS loss leads to unnatural strand orientations and unrealistic hairstyles (\cref{fig:4.4}).
Furthermore, the lack of \textcolor{black}{constrained multi-view images} prevents us from leveraging hair-specific features~\cite{hair_capture} to provide powerful supervision as in previous works~\cite{neural_strands}.

\begin{figure}[tp]
    \centering
    \includegraphics[width=0.45\textwidth]{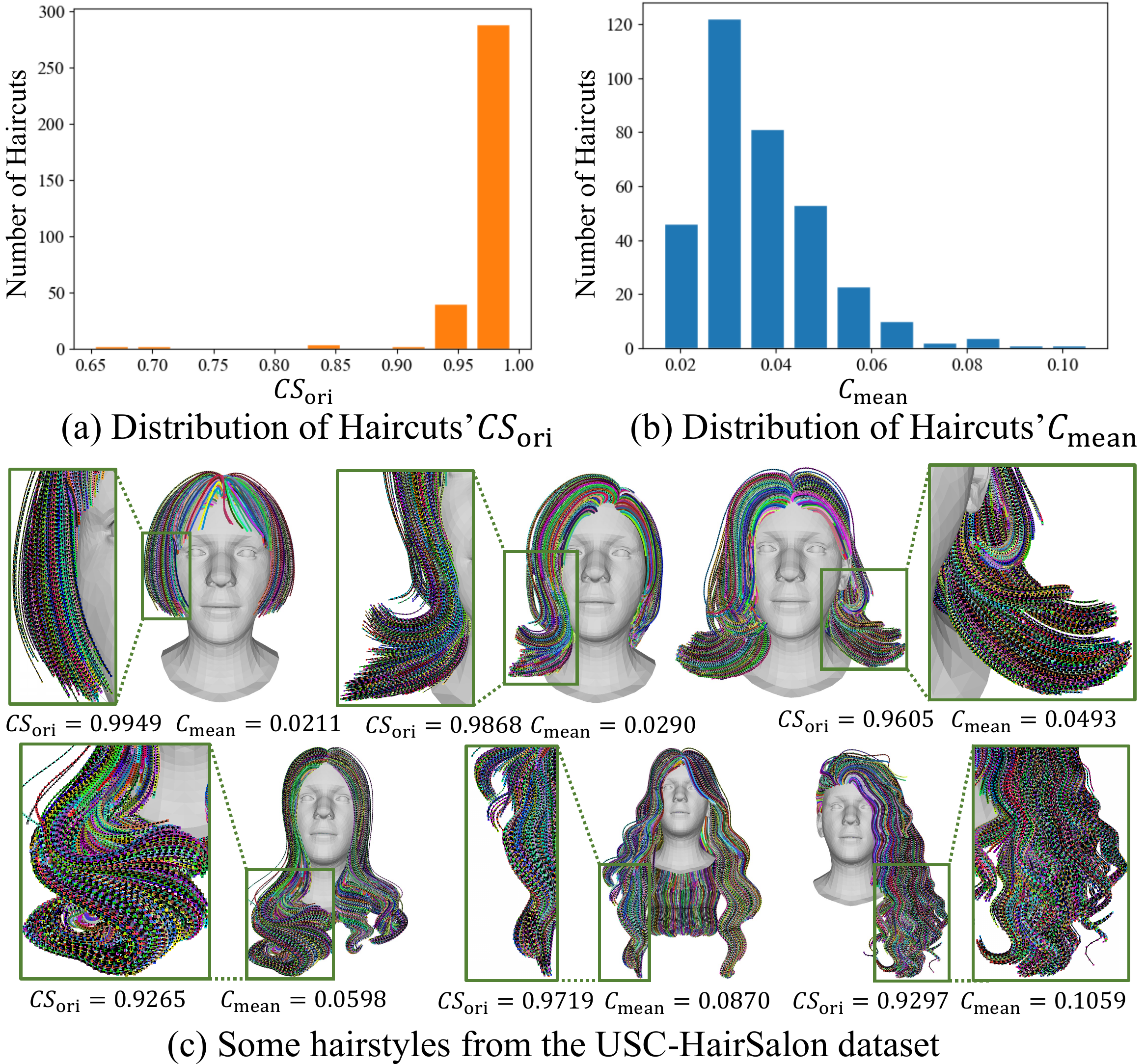} 
    \vspace{-0.3cm}
    \caption{Observation of hair geometric features: \textbf{(1) neighboring strand orientations are highly consistent.} \textbf{(2) strand curvature is strongly and positively related to the haircut curliness.}}
    \label{fig:3.3}
    \vspace{-0.5cm}
\end{figure}

\noindent
\textbf{Prior-Driven Losses.}
To address this issue, we propose two straightforward yet robust prior-driven losses based on observations of the 3D hair strand orientation and curvature to ensure the rationality of the generated hair strands.
Specifically, we observe two geometric properties of hair: \textbf{(1) neighboring strand orientations are highly consistent}, and \textbf{(2) strand curvature is strongly and positively correlated with the curliness of the hairstyle}. 
To validate the properties, we calculate the cosine similarity of adjacent hair strand orientations $CS_{\text{ori}}$ and the average curvature $C_{\text{mean}}$ of 343 hairstyles in the USC-HairSalon dataset~\cite{sing_view_rec_hair} using the following equations and visualize their distributions and some examples in \cref{fig:3.3}:
\vspace{-0.2cm}
\begin{align}
    CS_{\text{ori}} &= \frac{1}{N_{s}N_{p}} \sum_{i=1}^{N_{s}} \sum_{j=1}^{N_{p}} \sum_{k \in A(i)} \frac{o^{i}_{j} \cdot o^{k}_{j}}{|A(i)|},
    \label{eq:cs_ori} \\
    C_{\text{mean}} &= \frac{1}{N_{s}N_{p}}\sum^{N_{s}}_{i=1} \sum^{N_{p}}_{j=1} c^{i}_{j},
    \label{eq:c_mean}
    \vspace{-0.6cm}
\end{align}
where $A(i)$ is the set of adjacent strands for the $i$-th strand, and $|A(i)|$ is the number of neighboring strands.

As illustrated, over 95\% of hairstyles exhibit an orientation similarity above 0.9, and \textcolor{black}{the average curvature is significantly positively related to curliness,} which successfully validates the properties.
Based on these observations, we introduce orientation consistency loss and curvature regularization to guide the local and global strand shapes, formulated as follows:
\vspace{-0.2cm}
\begin{align}
    \mathcal{L}_{\text{ori}} &= 1 - CS_{\text{ori}},
    \label{eq:ori_cons} \\
    \mathcal{L}_{\text{cur}} &= \Vert C_{\text{mean}} - C_{\text{target}} \Vert_{1},
    \label{eq:cur_reg}
    \vspace{-0.5cm}
\end{align}
where $C_{\text{target}}$ represents the target average curvature set according to the input hair description.
These two losses, inspired by 3D hair geometry priors, regularize the hair shape by supervising the consistency of orientations between adjacent strands and the overall hairstyle's curvature.

\begin{figure*}[tp]
    \centering
    \includegraphics[width=0.88\textwidth]{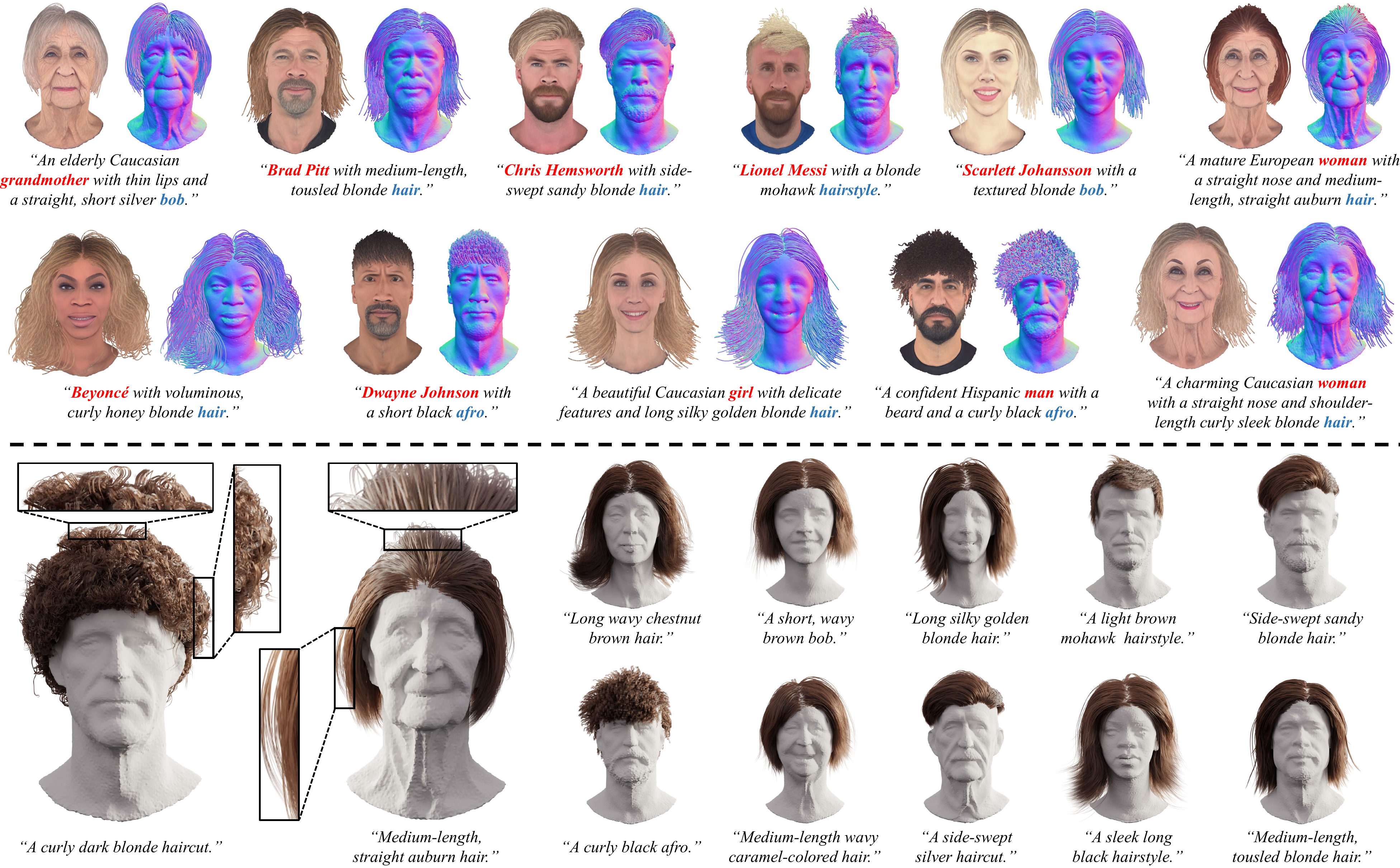} 
    \vspace{-0.3cm}
    \caption{Examples of high-fidelity and diverse 3D heads and strand-accurate haircuts generated by our method.
    The upper visualization includes rendered color and normal maps of the head and hair prismatic meshes. 
    The lower visualization shows the physics-based hair strand rendering result using Blender~\cite{blender}. For better strand-based visualization, we interpolate generated hair to approximately 10,000 strands and apply a consistent appearance.
    \textbf{Please zoom in for detailed views, and refer to the Supp. Mat. for video demonstrations.}}
    \label{fig:4.1}
    \vspace{-0.55cm}
\end{figure*}

Additionally, we introduce a series of losses $\mathcal{L}_{bbox}$, $\mathcal{L}_{face}$, and $\mathcal{L}_{colli}$, which prevent the hair from exceeding the bounding box, obscuring the face, and colliding with the head, respectively.
These losses can further improve the rationality of hair geometry. 
Their details are provided in the Supp. Mat. 
The final loss function for optimizing hair shape is expressed as follows:
\vspace{-0.1cm}
\begin{align}
    \mathcal{L}_{\text{hair-geo}} =\ & \mathcal{L}^{\text{hd}}_{SDS} + \lambda^{\text{hn}}_{SDS} \mathcal{L}^{\text{hn}}_{SDS} + \lambda_{\text{ori}} \mathcal{L}_{\text{ori}} + \lambda_{\text{cur}} \mathcal{L}_{\text{cur}} \nonumber \\
    & + \lambda_{\text{bbox}} \mathcal{L}_{\text{bbox}} + \lambda_{\text{face}} \mathcal{L}_{\text{face}} + \lambda_{\text{colli}} \mathcal{L}_{\text{colli}}.
\label{eq:hair_geometry}
\end{align}

\subsection{Hair Texture Generation}
\label{sec:3.5}
\noindent
Next, we fix the optimized hair strand geometry and model the realistic hair strand texture under the supervision of the SDS loss as follows (\cref{fig:3.1}-(c)):
\vspace{-0.1cm}
\begin{equation}
    \nabla_{{\psi}_\text{s}} \mathcal{L}^{\text{hc}}_{SDS} = \mathbb{E}_{t, \epsilon} ( {\epsilon}_{{\phi}_\text{hc}}(c^\text{h+s}_t; n^\text{h+s}, y_\text{h+s}, t) - \epsilon ) \frac{\partial c^\text{h+s}}{\partial {\psi}_\text{s}},
    \label{eq:sds_hair_color}
    \vspace{-0.1cm}
\end{equation}
where ${\phi}_\text{hc}$ is the human-specific normal-conditioned diffusion model from HumanNorm~\cite{humannorm}, $c^\text{h+s}$ is the rendered color map of the head texture field ${\psi}_\text{h}$ and hair strand texture field ${\psi}_\text{s}$.
Since the vanilla SDS loss often leads to color oversaturation, we replace it with the following MSDS loss~\cite{humannorm} to further enhance the texture's realism in later iterations of texture optimization:
\vspace{-0.1cm}
\begin{equation}
    \nabla_{{\psi}_\text{s}} \mathcal{L}^{\text{hc}}_{MSDS} = \mathbb{E}_{t, \epsilon} ( H(c^\text{h+s}_t; n^\text{h+s}, y_\text{h+s}, t) - \epsilon ) \frac{\partial c^\text{h+s}}{\partial {\psi}_\text{s}},
    \label{eq:msds_hair_color}
    \vspace{-0.1cm}
\end{equation}
where $H(\cdot)$ denotes the multi-step operation function.
Additionally, we propose a strand-aware texture field that models orientation-dependent texture to better generate high-frequency color variations. Please refer to the Supp. Mat. for more their details.

\subsection{Implementation Details}
\label{sec:3.6}
Our method is implemented using PyTorch~\cite{pytorch} and ThreeStudio~\cite{threestudio}. Experiments are conducted on an Ubuntu server equipped with A6000 GPUs.
Starting with a prompt such as ``A man with brown hair'', we first generate a bald head using ``A bald man''. Then, we generate the hair based on ``A man with brown hair''.
Further implementation details are available in the Supp. Mat.
\begin{figure*}[tp]
    \centering
    \includegraphics[width=0.9\textwidth]{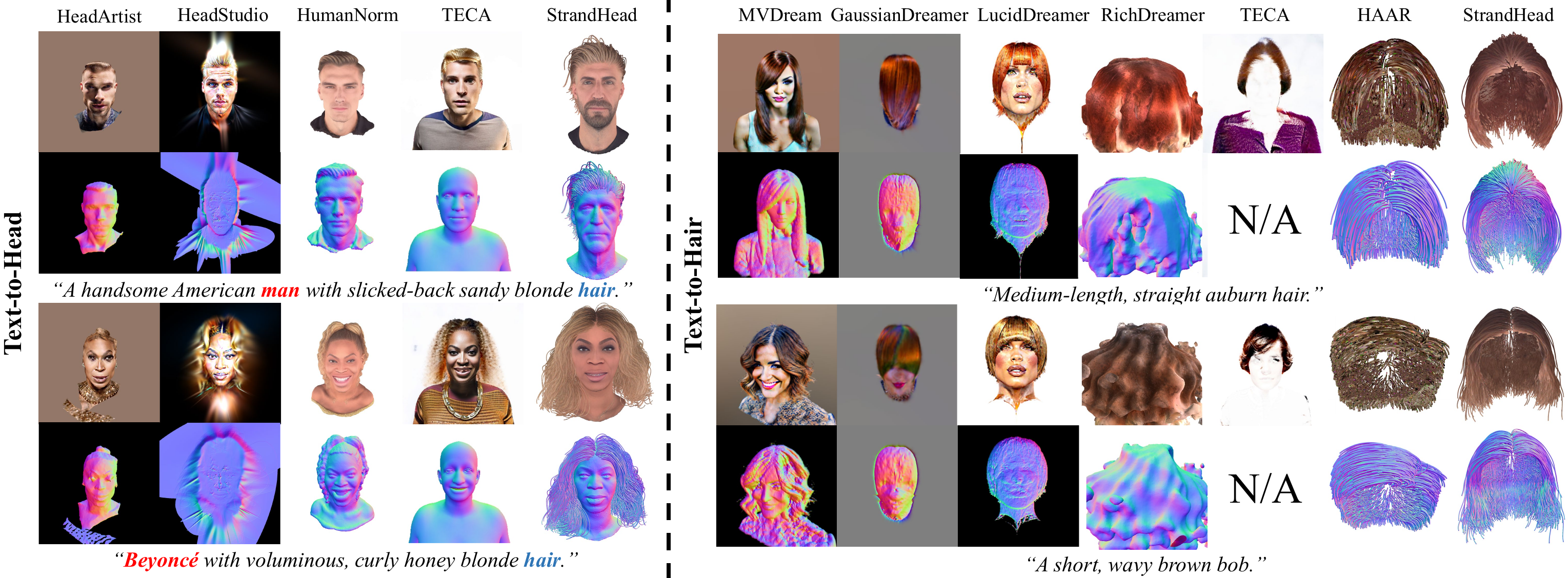} 
    \vspace{-0.3cm}
    \caption{Qualitative comparisons with the SOTA methods. Since TECA~\cite{teca} uses the vanilla NeRF to represent hair, rendering normals is not supported. HAAR~\cite{haar} generates only the geometry of hair strands, so we first convert the strands into prismatic meshes using differentiable prismatization and then utilize TEXTure~\cite{texture} to generate texture for visualization and comparison.}
    \label{fig:4.2}
    \vspace{-0.5cm}
\end{figure*}

\begin{figure}[tp]
    \centering
    \includegraphics[width=0.4\textwidth]{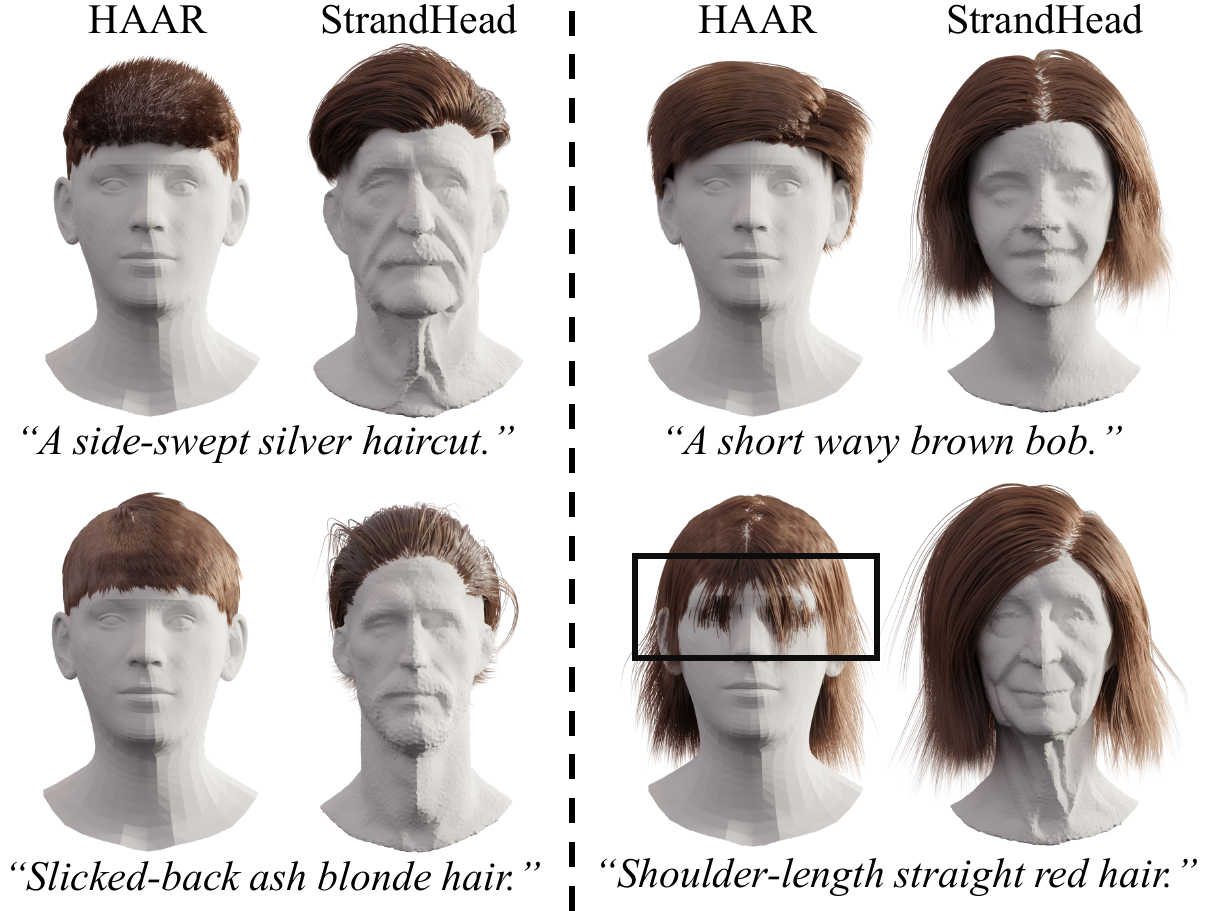}
    \vspace{-0.3cm}
    \caption{Qualitative comparison with HAAR~\cite{haar}. \textbf{HAAR, which does not model heads, often produces unreasonable hair-head collisions (highlighted in the black box).}}
    \label{fig:new4.3}
    \vspace{-0.5cm}
\end{figure}

\section{Experiments}
\label{sec:4}
Some examples of 3D head models with strand-based hair generated by StrandHead are shown in \cref{fig:4.1}. 
Due to space limitations, only the key experimental settings and results are presented here. 
For more results, evaluations, and details, please refer to our Supp. Mat.

\subsection{Experimental Settings}
\label{sec:4.1}

\noindent
\textbf{Baselines.}
We compare StrandHead with the SOTA methods for head avatar and haircut generation. 
The baseline methods for head generation include text-to-holistic-head works (HeadArtist~\cite{headartist}, HumanNorm~\cite{humannorm}, and HeadStudio~\cite{headstudio}) and text-to-decoupled-head work (TECA~\cite{teca}).
For text-to-hair generation, we consider general text-to-3D methods (MVDream~\cite{mvdream}, GaussianDreamer~\cite{gaussiandreamer}, LucidDreamer~\cite{luciddreamer}, and RichDreamer~\cite{richdreamer}), text-to-strand method (HAAR~\cite{haar}), and TECA~\cite{teca}.

\noindent
\textbf{Dataset Construction.}
To create the dataset, we use ChatGPT\cite{chatgpt} to randomly generate 30 text descriptions of heads with hair.
These include 30 descriptions for assessing head generation and 30 for evaluating hair generation.
The full list of prompts can be found in the Supp. Mat.

\noindent
\textbf{Evaluation Metrics.}
Current text-to-3D methods typically use CLIP-based metrics to assess text-image alignment and output quality.
However, research~\cite{t2i-compbench,llmscore} shows that these metrics struggle to capture fine-grained alignment between 3D content and input prompts, as validated by our experiments in the Supp. Mat.
To address this limitation, we draw inspiration from Progressive3D~\cite{progressive3d} and Barbie~\cite{barbie}, using BLIP-VQA~\cite{blip,lavis} and BLIP2-VQA~\cite{blip2,lavis} for evaluation.
Specifically, we convert each prompt into two questions to separately verify head and hair. 
We then feed the rendered image of the generated 3D content into the VQA model, asking each question in sequence and using the probability of a ``yes''  response as our evaluation metric. 
For example, the head prompt ``A man with black hair'' is transformed into ``Is the person in the picture a man?'' and ``Does the person in the picture have black hair?'' The hair prompt ``Black hair'' becomes ``Is the object in the picture black hair?''
Finally, we conduct a user study by randomly selecting 10 generated examples and asking 30 volunteers to assess \textbf{(1)} generation quality and \textbf{(2)} text-image alignment, and select the best methods.

\begin{table}[tp]
    \centering
    \resizebox{0.8\linewidth}{!}{
        \begin{tabular}{ccccc}
            \toprule
            Method & BLIP-VQA $\uparrow$ & BLIP2-VQA $\uparrow$ & GQP $\uparrow$ & TAP $\uparrow$ \\
            \midrule
            HeadArtist & 0.7667 & \textbf{0.9667} & 1.00 & 2.33 \\
            HeadStudio & \underline{0.7833} & 0.8833 & 3.33 & 3.67 \\
            HumanNorm & 0.7000 & \underline{0.9500} & 7.67 & 7.67 \\
            TECA & 0.7333 & \underline{0.9500} & \underline{34.33} & \underline{28.33} \\
            StrandHead (Ours) & \textbf{0.8500} & \textbf{0.9667} & \textbf{53.67} & \textbf{58.00} \\
            \midrule
            MVDream & \textbf{0.9000} & 0.8333 & \underline{24.67} & \underline{20.00} \\
            GaussianDreamer & 0.3333 & 0.3667 & 5.33 & 3.00 \\
            LucidDreamer & 0.8000 & \textbf{0.9333} & 5.33 & 5.00 \\
            RichDreamer & \underline{0.8333} & 0.7667 & 4.00 & 5.67 \\
            TECA & 0.7000 & 0.7667 & 1.67 & 3.67 \\
            HAAR & 0.6333 & 0.2000 & 1.33 & 2.33 \\
            StrandHead (Ours) & \textbf{0.9000} & \underline{0.9000} & \textbf{57.67} & \textbf{60.33} \\
            \bottomrule
        \end{tabular}
    }
    \vspace{-0.3cm}
    \caption{Quantitative comparisons with the SOTA methods. The best and second-best results are highlighted in \textbf{bolded} and \underline{underlined}, respectively. GQP: generation quality preference (\%), TAP: text-image alignment preference (\%).} 
    \label{tab:quantitative_comparisons}
    \vspace{-0.6cm}
\end{table}

\subsection{Comparisons of Head Generation}
\label{sec:4.2}
As illustrated in the upper section of \cref{tab:quantitative_comparisons}, StrandHead outperforms all comparison methods across every evaluation metric.
The qualitative results on the left side of \cref{fig:4.2} further highlight the advantages of our method.
Compared to other head generation methods, the head avatars generated by StrandHead exhibit not only more refined facial geometry and texture details but also strand-accurate hair with a plausible appearance that integrates seamlessly with physics-based simulation systems.
To the best of our knowledge, StrandHead is the first head generation framework that realizes strand-level hair modeling, a capability that holds substantial potential for advancing human-centric 3D AIGC applications in the industry.

\subsection{Comparisons of Hair Generation}
\label{sec:4.3}
As shown in the lower part of \cref{tab:quantitative_comparisons}, our approach surpasses other methods across most evaluation metrics, and is only slightly lower than LucidDreamer (the latest open-source text-to-3D method) on BLIP2-VQA, ranking the second. 
The qualitative comparisons on the right side of \cref{fig:4.2} display sample generation results.
Compared to general text-to-3D methods, our approach more precisely captures the internal structure of hair strands, resulting in realistic haircut geometry and appearance without incorporating incorrect content such as parts of the human head.

Compared with HAAR~\cite{haar}, StrandHead relies on robust 2D/3D human-centric priors rather than large-scale paired datasets to supervise hair generation. 
This not only allows it to generate hairstyles that are uncommon in the dataset (see slicked-back and side-swept hairstyles in \cref{fig:new4.3}), but also models hair texture or geometry that adapts to specific head shapes (\cref{fig:4.3}-(b)), thereby avoiding unnatural hair-head collisions (see the black box in \cref{fig:new4.3}).

\subsection{Ablation Study}
\label{sec:4.4}

\begin{figure}[tp]
    \centering
    \includegraphics[width=0.4\textwidth]{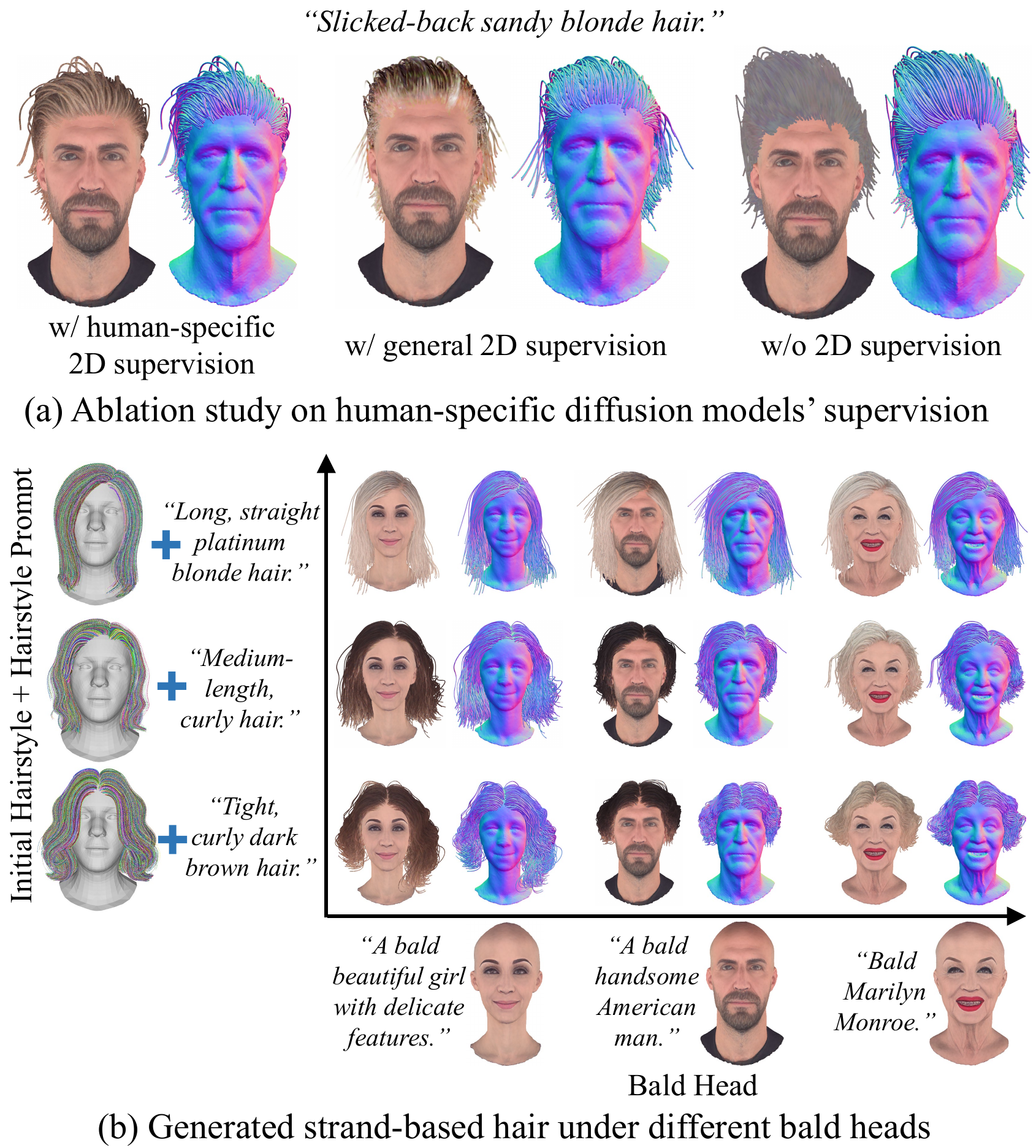} 
    \vspace{-0.3cm}
    \caption{Analysis of human-specific 2D diffusion models.}
    \label{fig:4.3}
    \vspace{-0.5cm}
\end{figure}

\noindent
\textbf{Effectiveness of Human-Specific 2D Generative Priors.}
We demonstrate the importance of human-specific 2D generative priors from two aspects:
\textbf{(1)} As shown in \cref{fig:4.3}-(a), ignoring 2D supervision fails to generate meaningful hair. While a general 2D diffusion model can slightly improve modeling performance, it still falls short of producing high-quality results. Only by incorporating human-specific 2D generative priors can one achieve realistic textures and reasonable shapes for hairstyles.
\textbf{(2)} \cref{fig:4.3}-(b) illustrates generated hair under varying bald head conditions while maintaining a fixed initial hairstyle and hair prompt. The hairstyles created under the supervision of human-specific 2D diffusion models exhibit clear geometric and textural variations that adapt to specific bald heads. This strongly demonstrates the necessity of considering the bald head for hair strand generation and highlights the advantages of StrandHead over HAAR~\cite{haar}.

\begin{figure}[tp]
    \centering
    \includegraphics[width=0.40\textwidth]{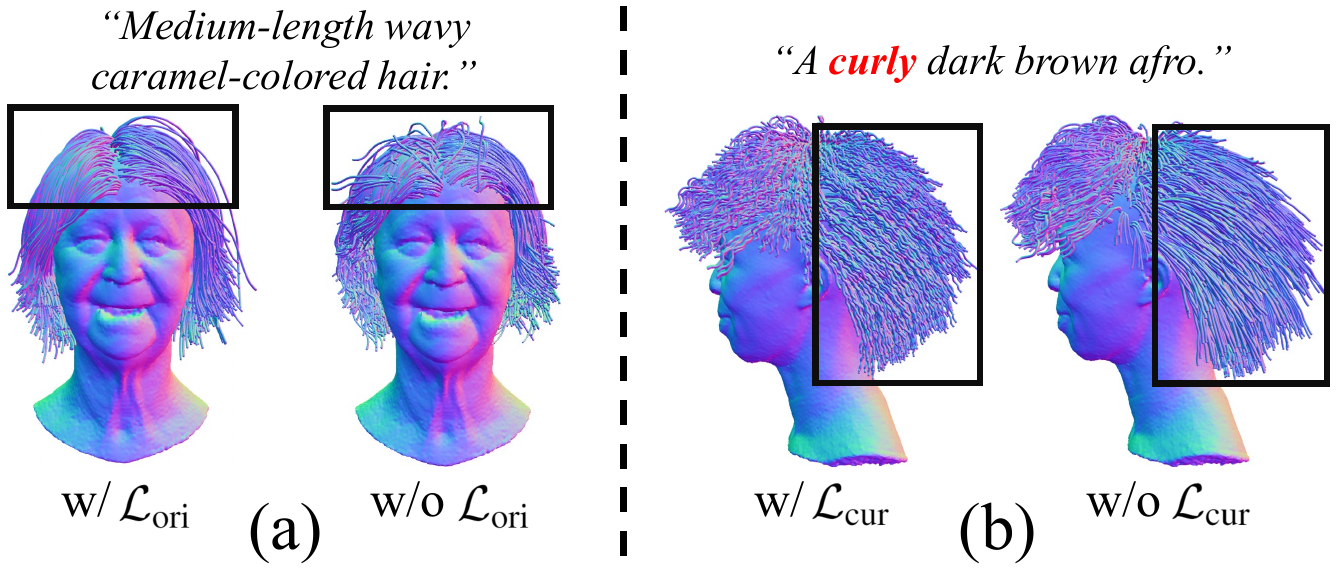} 
    \vspace{-0.3cm}
    \caption{Ablation study on \textbf{(a)} orientation consistency loss $\mathcal{L}_{\text{ori}}$ and \textbf{(b)} curvature regularization loss $\mathcal{L}_{\text{cur}}$.}
    \label{fig:4.4}
    \vspace{-0.5cm}
\end{figure}

\noindent
\textbf{Effectiveness of Prior-Driven Losses.}
\cref{fig:4.4}-(a) demonstrates that incorporating $\mathcal{L}_{\text{ori}}$ provides strong guidance for generating natural hair by supervising the consistency of orientations between adjacent strands.
As shown in \cref{fig:4.4}-(b), our $\mathcal{L}_{\text{cur}}$ preserves the desired curliness by constraining the overall hairstyle's curvature.
In summary, these two losses, inspired by 3D hair geometry priors, effectively enhance the realism and rationality of the generated haircut.




\begin{figure}[tp]
    \centering
    \includegraphics[width=0.45\textwidth]{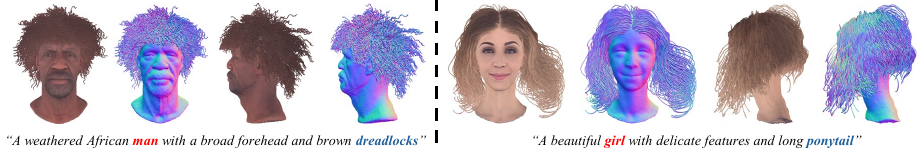} 
    \vspace{-0.3cm}
    \caption{Failure cases.}
    \label{fig:4.5}
    \vspace{-0.7cm}
\end{figure}

\section{Conclusion}
\label{sec:5}
In this paper, a novel framework StrandHead is proposed to generate realistic hair-disentangled 3D head avatars using text prompts. With a series of 2D generative models pre-trained on human-centric data, StrandHead generates strand-based hair requiring no large-scale text-hair paired data for supervision. This goal is achieved mainly by two approaches: \textbf{(1)} proposing a novel differentiable prismatization method that transforms the hair strands into prismatic meshes, and \textbf{(2)} using the prior-driven losses inspired by the observation of real-world haircut and the statistical features of human-crafted 3D hair data.
The former one produces watertight geometry more similar to the original hair strands, boosting a stable optimization with significantly less normal ambiguity. 
Besides, the latter one well constrains the optimization with reliable guidance to generate reasonable 3D haircuts. Extensive experiments demonstrate that StrandHead obtains the state-of-the-art performance on 3D head avatars and haircuts generation. The generated results are easily implemented in industrial software to produce physical simulation and high-fidelity rendering.

\noindent
\textbf{Limitations and Future Work.}
Due to the limited representation capabilities of the strand generator, StrandHead is unable to create highly complex 3D hairstyles (e.g., dreadlocks and ponytails in \cref{fig:4.5}). Additionally, although the SDS-based method effectively leverages 2D generative priors, its high computational cost restricts practical application. In future work, we aim to address these challenges by incorporating richer hairstyle datasets and exploring deeper hair priors.

\noindent
\textbf{Acknowledgement.}
This work was supported by the National Science Foundation of China (62376121).

{
    \small
    \bibliographystyle{ieeenat_fullname}
    \bibliography{main}
}
\clearpage
\setcounter{page}{1}

\twocolumn[{
\renewcommand\twocolumn[1][]{#1}
\maketitlesupplementary
\vspace{-3mm}
\begin{center}
    \captionsetup{type=figure}
    \includegraphics[width=0.9\textwidth]{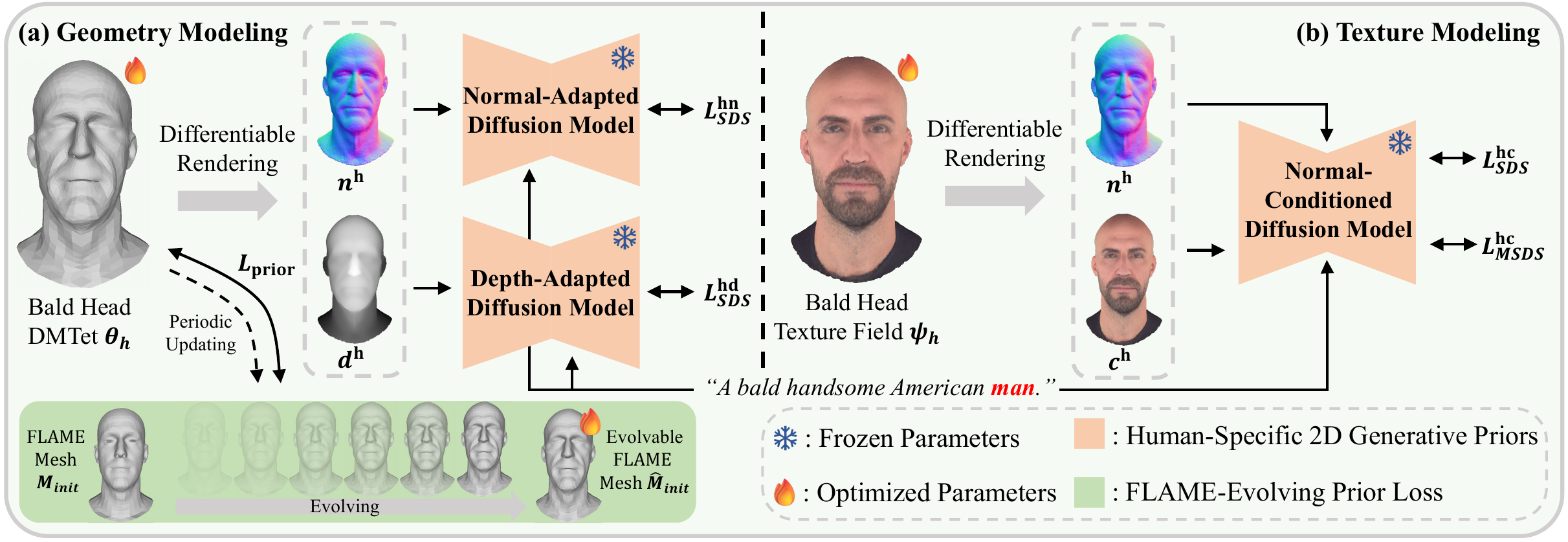} 
    \vspace{-0.3cm}
    \caption{The process for generating a bald head model involves two steps:
    \textbf{(a)} Employing human-specific geometry-aware diffusion models and FLAME-evolving prior loss to model realistic and semantic-aligned bald head shapes.
    \textbf{(b)} Subsequently, using a normal-conditioned diffusion model to generate lifelike head textures.}
    \label{fig:pipeline}
\end{center}
}]

\section{Supplementary Material}
This document includes the following supplementary content:
\begin{itemize}
    \item Bald Head Generation.
    \item Implementation Details.
    \item More Evaluations.
    \item More Experiment Details and Results.
    \item Prompt List.
    \item Ethics Statement.
\end{itemize}

\subsection{Bald Head Generation}
\noindent
Following HumanNorm~\cite{humannorm} and HeadArtist~\cite{headartist}, 
We use DMTet~\cite{dmtet} ${\theta}_\text{h}$ and a texture field ${\psi}_\text{h}$ to model the geometry and appearance of the bald head respectively. These components are optimized separately in two stages, as illustrated in \cref{fig:pipeline}.

\noindent
\textbf{Head Geometry Modeling.}
Specifically, we first initialize the head DMTet ${\theta}_\text{h}$ utilizing the FLAME model $M_\text{init}$. We then refine it under the supervision of human-specific geometry-aware diffusion models (\cref{fig:pipeline}-(a)) by leveraging the following losses:
\vspace{-0.1cm}
\begin{align}
    \nabla_{{\theta}_\text{h}} \mathcal{L}^{\text{hn}}_{SDS} &= \mathbb{E}_{t, \epsilon} \left[( {\epsilon}_{{\phi}_\text{hn}}(n^\text{h}_t; y_\text{h}, t) - \epsilon ) \frac{\partial n^\text{h}}{\partial {\theta}_\text{h}} \right],
    \label{eq:sds_head_normal} \\[-0.1cm]
    \nabla_{{\theta}_\text{h}} \mathcal{L}^{\text{hd}}_{SDS} &= \mathbb{E}_{t, \epsilon} \left[( {\epsilon}_{{\phi}_\text{hd}}(d^\text{h}_t; y_\text{h}, t) - \epsilon ) \frac{\partial d^\text{h}}{\partial {\theta}_\text{h}} \right],
    \label{eq:sds_head_dpeth}
    \vspace{-0.5cm}
\end{align}
where $y_\text{h}$ is the input bald head text, ${\phi}_\text{hn}$ and ${\phi}_\text{hd}$ are the human-specialized normal-adapted and depth-adapted diffusion models, respectively, and $n^\text{h}$ and $d^\text{h}$ are the rendered normal and depth maps of the bald head.

In addition, to obtain semantic information for accurate hair initialization, we introduce a FLAME-evolving prior loss, inspired by Barbie~\cite{barbie}.
In specific, we freeze the FLAME shape parameters $\beta$ but improve $M_{\text{init}}$ to evolvable $\hat{M}_{\text{init}}$ by adding learnable vertex-wise offsets. We periodically fit $\hat{M}_{\text{init}}$ to the head DMTet every $\delta$ iteration, thereby obtaining accurate semantic information.
Moreover, $\hat{M}_\text{init}$ provide a reliable and diverse head prior to prevent unnatural geometry using the following formula:
\vspace{-0.1cm}
\begin{equation}
    \mathcal{L}_{\text{prior}} = \sum_{p \in P} \left\| s_{\theta_\text{h}}(p) - s_{\hat{M}_{\text{init}}}(p) \right\|_2^2,
    \label{eq:prior}
    \vspace{-0.1cm}
\end{equation}
where $s_{\theta_\text{h}}$ and $s_{\hat{M}_{\text{init}}}$ are the signed distance functions (SDF) of the head DMTet ${\theta}_\text{h}$ and $\hat{M}_{\text{init}}$, respectively, and $P$ is a set of randomly sampled points. 
Through this evolving process, $\hat{M}_{\text{init}}$ gradually captures rich geometric features (\eg, beards and wrinkles), providing reliable yet diverse priors for subsequent geometry generation (\cref{fig:flame_loss}).
In summary, the loss function for optimizing the head geometry is as follows:
\vspace{-0.1cm}
\begin{equation}
\mathcal{L}_{\text{head-geo}} = \mathcal{L}^{\text{hn}}_{SDS} + \mathcal{L}^{\text{hd}}_{SDS} + \lambda_{\text{prior}} \mathcal{L}_{\text{prior}}.
\label{eq:head_geometry}
\vspace{-0.1cm}
\end{equation}

\noindent
\textbf{Head Texture Modeling.}
Given the generated head geometry generated, we fix it and utilize a texture field ${\psi}_\text{h}$, which maps a query position to its color to generate head appearance.
Specifically, we construct this field using MLP with multi-resolution hash encoding~\cite{instant-ngp}, and we optimize it using the following loss function (\cref{fig:pipeline}-(b)):
\vspace{-0.2cm}
\begin{equation}
    \nabla_{{\psi}_\text{h}} \mathcal{L}^{\text{hc}}_{SDS} = \mathbb{E}_{t, \epsilon} \left[( {\epsilon}_{{\phi}_\text{hc}}(c^\text{h}_t; n^\text{h}, y_\text{h}, t) - \epsilon ) \frac{\partial c^\text{h}}{\partial {\psi}_\text{h}} \right],
    \label{eq:sds_head_color}
    \vspace{-0.2cm}
\end{equation}
where ${\phi}_\text{hc}$ is a human-specialized normal-conditioned diffusion model, and $c^\text{h}$ represents the rendered color image of the generated head.
Since the vanilla SDS loss often leads to color oversaturation, we replace it with the following MSDS loss~\cite{humannorm} to further enhance the texture's realism in later iterations of texture optimization:
\begin{equation}
    \begin{aligned}
    &\nabla_{{\psi}_\text{h}} \mathcal{L}^{\text{hc}}_{MSDS} = \mathbb{E}_{t, \epsilon} \left[( h(c^\text{h}_t; n^\text{h}, y_\text{h}, t) - \epsilon ) \frac{\partial c^\text{h}}{\partial {\psi}_\text{h}} \right] + \\ & \mathbb{E}_{t, \epsilon} \left[( V(h(c^\text{h}_t; n^\text{h}, y_\text{h}, t)) - V(\epsilon) ) \frac{\partial V(\epsilon)}{\partial \epsilon} \frac{\partial c^\text{h}}{\partial {\psi}_\text{h}} \right],
    \label{eq:msds_head_color}
    \vspace{-0.2cm}
    \end{aligned}
\end{equation}
where $V$ is the first $k$ layers of the VGG network~\cite{vgg}
$h(\cdot)$ denotes the multi-step image generation function of the normal-aligned diffusion model.

\subsection{Implementation Details}
\textbf{Hyper-Parameters.}
For the bald head generation, our approach requires 10,000 iterations for geometry creation, 2,000 iterations for texture synthesis, and 5,000 iterations for appearance refinement using MSDS~\cite{humannorm}. The loss weight $\lambda_{\text{prior}}$ is set to $1 \times 10^{3}$.
For hair generation, the process involves 5,000 iterations for geometry modeling, 2,000 iterations for texture generation, and 5,000 iterations for visual enhancement using MSDS~\cite{humannorm}. The respective loss weights for $\lambda^{\text{hn}}_{SDS}$, $\lambda_{\text{ori}}$, $\lambda_{\text{cur}}$, $\lambda_{\text{bbox}}$, $\lambda_{\text{face}}$, and $\lambda_{\text{colli}}$ are assigned as $1 \times 10^{-3}$, $1 \times 10^{4}$, $1 \times 10^{4}$, $1 \times 10^{3}$, $1 \times 10^{3}$, and $1 \times 10^{3}$ respectively.
The number of strand polylines $N_{s}$ and strand points $N_{p}$ are configured as 3000 and 100, respectively. 
For different hairstyles—normal, straight, wavy, and curly—the target curvature $C_{\text{target}}$ is defined as $5 \times 10^{-2}$, $2 \times 10^{-2}$, $1 \times 10^{-1}$, and $2 \times 10^{-1}$, respectively.
All experiments are conducted on an Ubuntu server equipped with A6000 GPUs.
Generating a bald head takes roughly 4 hours with 24GB of memory, and generating a haircut takes roughly 4 hours with 44GB of memory.

\begin{figure}[tp]
    \centering
    \includegraphics[width=0.43\textwidth]{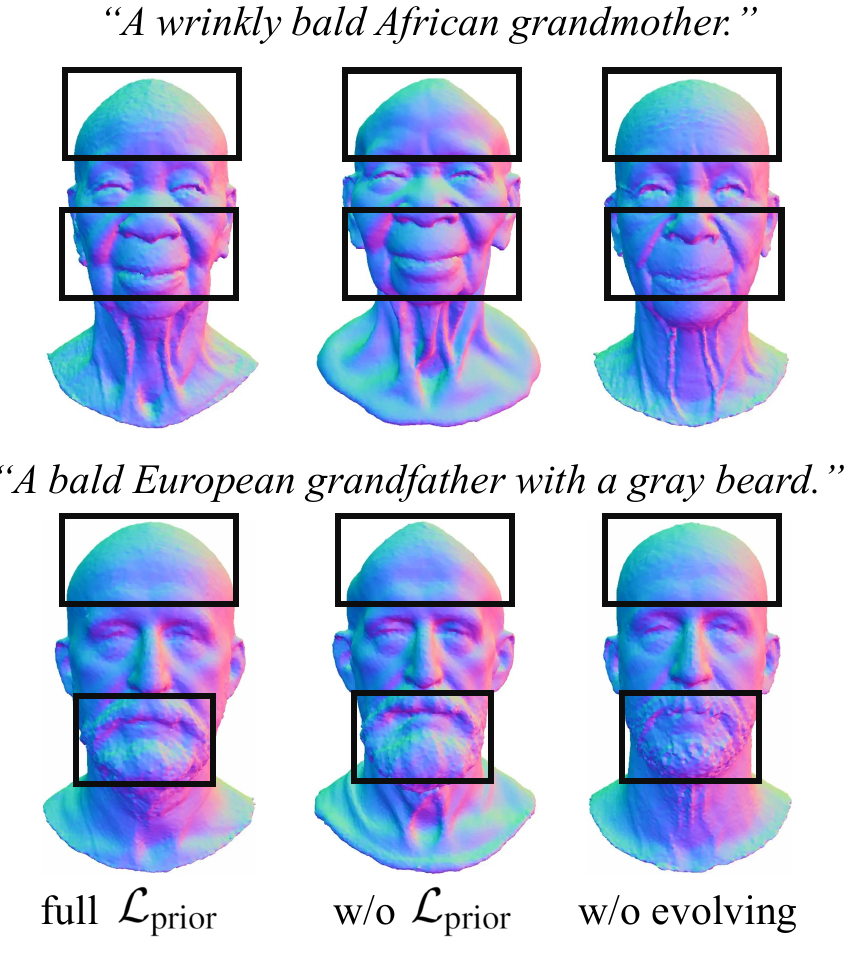} 
    \vspace{-0.2cm}
    \caption{Ablation study on FLAME-evolving prior loss.}
    \label{fig:flame_loss}
    \vspace{-0.2cm}
\end{figure}

\noindent
\textbf{Details of FLAME-Evolving Prior Loss.}
In the FLAME-evolving prior loss, we periodically optimize $\hat{M}_\text{init}$ to fit the current human head geometry at every 1000 iterations, thus providing an effective and flexible human head prior constraint.
The fitting loss function is as follows:
\begin{equation}
    \mathcal{L}_\text{fit} = \lambda_{\text{chamf}} \mathcal{L}_\text{chamf} + \lambda_{\text{edge}} \mathcal{L}_\text{edge} + \lambda_{\text{nor}} \mathcal{L}_\text{nor} + \lambda_{\text{lap}} \mathcal{L}_\text{lap},
\end{equation}
where $\mathcal{L}_\text{chamf}$ is the Chamfer distance between $\hat{M}_\text{init}$ and the current human head geometry, $\mathcal{L}_\text{edge}$ is the edge length regularization loss, $\mathcal{L}_\text{nor}$ is the normal consistency loss, and $\mathcal{L}_\text{lap}$ is the Laplacian smoothness loss.
The loss weights $\lambda_{\text{chamf}}$, $\lambda_{\text{edge}}$, $\lambda_{\text{nor}}$, and $\lambda_{\text{lap}}$ are set to $1 \times 10^{2}$, $1 \times 10^{0}$, $1 \times 10^{-2}$, and $1 \times 10^{-1}$, respectively.
The ablation study results for the FLAME-evolving prior loss are presented in \cref{fig:flame_loss}. 
As shown, excluding $\mathcal{L}_\text{prior}$ results in exaggerated head shapes (\eg, overly pointed head tops), significantly compromising the realism of the outputs. 
Employing a non-evolving prior loss achieves reasonable head proportions but fails to preserve finer geometric details such as beards and wrinkles. 
By leveraging the full $\mathcal{L}_\text{prior}$, our approach generates head geometry with both accurate proportions and high detail fidelity.

\begin{figure}[tp]
    \centering
    \includegraphics[width=0.43\textwidth]{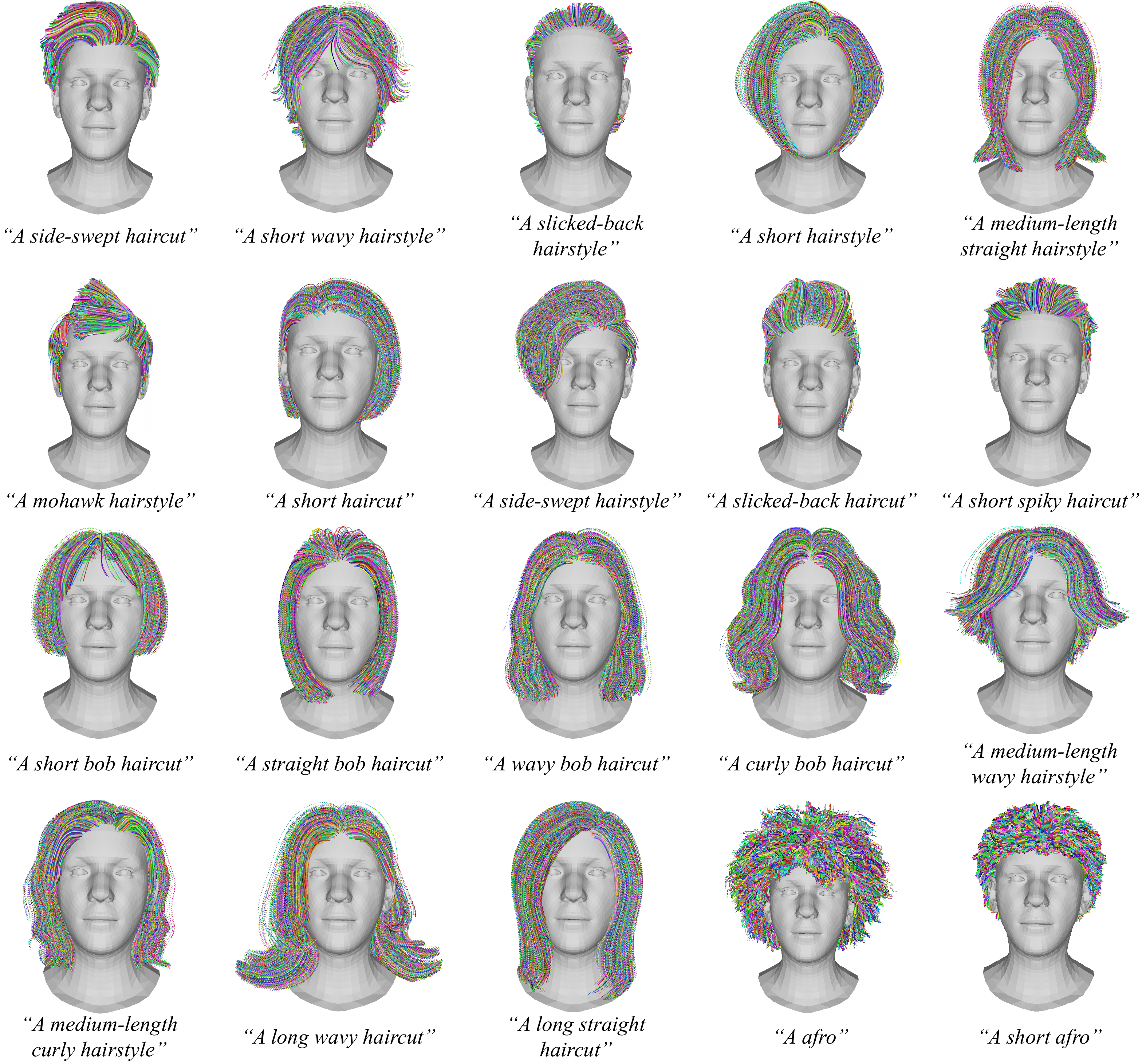} 
    \vspace{-0.3cm}
    \caption{Initialize hairstyles. Since the USC-HairSalon Dataset~\cite{sing_view_rec_hair} lacks afro hairstyles, we use HAAR~\cite{haar} to generate the initial afro hairstyle.}
    \label{fig:init_hair}
    \vspace{-0.3cm}
\end{figure}

\noindent
\textbf{Details of Hair Initialization.}
As mentioned in the main paper, we utilize ChatGPT\cite{chatgpt} to select the most representative hairstyles from the USC-HairSalon Dataset \cite{sing_view_rec_hair}.
Specifically, we first exclude some exaggerated hairstyles (e.g., those that are overly long or excessively messy).
Next, ChatGPT is used to generate textual descriptions for each hairstyle, which are then utilized to select the 20 most representative ones.
As shown in \cref{fig:init_hair}, the selected hairstyles cover different lengths, curvatures, and styles, ensuring rich diversity.
Finally, we optimize neural scalp textures (NST) to fit the selected hairstyle using \cref{eq:init_hair}, where $\lambda_{\text{ori}}$ and $\lambda_{\text{cur}}$ are set $5 \times 10^{-2}$, and $1 \times 10^{0}$, respectively.

\begin{figure}[tp]
    \centering
    \includegraphics[width=0.45\textwidth]{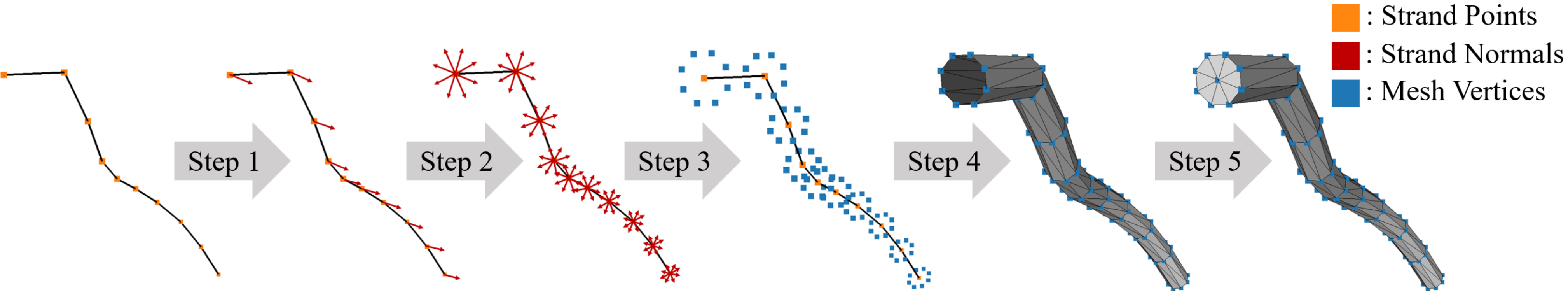} 
    \vspace{-0.3cm}
    \caption{Illustration of converting a hair strand into a prismatic mesh using the differentiable prismatization algorithm.}
    \label{fig:dp_pipeline}
    \vspace{-0.3cm}
\end{figure}

\begin{figure}[tp]
    \centering
    \includegraphics[width=0.45\textwidth]{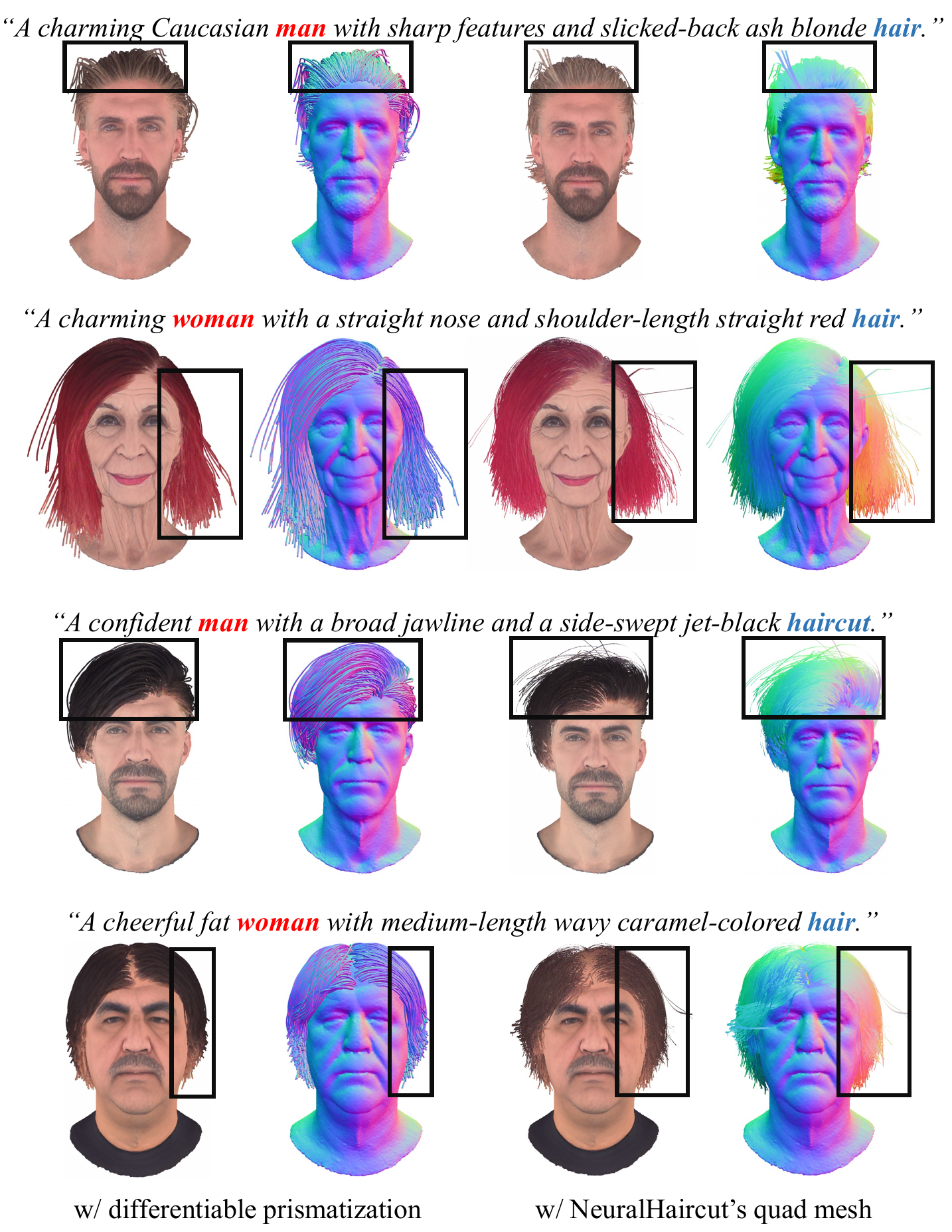} 
    \vspace{-0.3cm}
    \caption{Differentiable prismatization vs. NeuralHaircut's quad mesh~\cite{neural_haircut}.}
    \label{fig:dp}
    \vspace{-0.3cm}
\end{figure}

\begin{figure}[tp]
    \centering
    \includegraphics[width=0.45\textwidth]{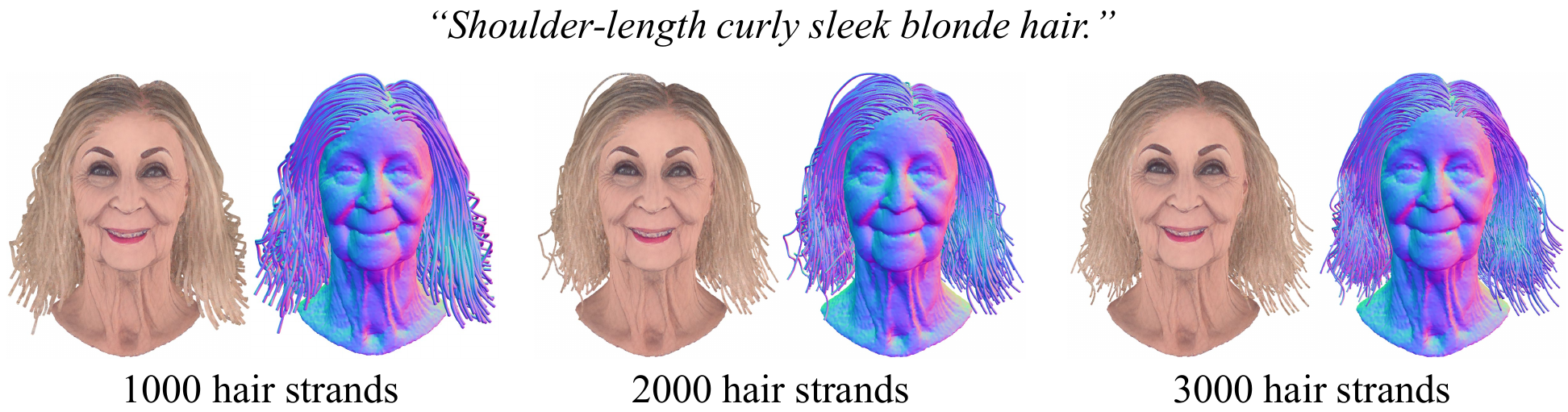} 
    \vspace{-0.3cm}
    \caption{Ablation study on the number of hair strands.}
    \label{fig:strand_number}
    \vspace{-0.3cm}
\end{figure}

\begin{figure}[tp]
    \centering
    \includegraphics[width=0.45\textwidth]{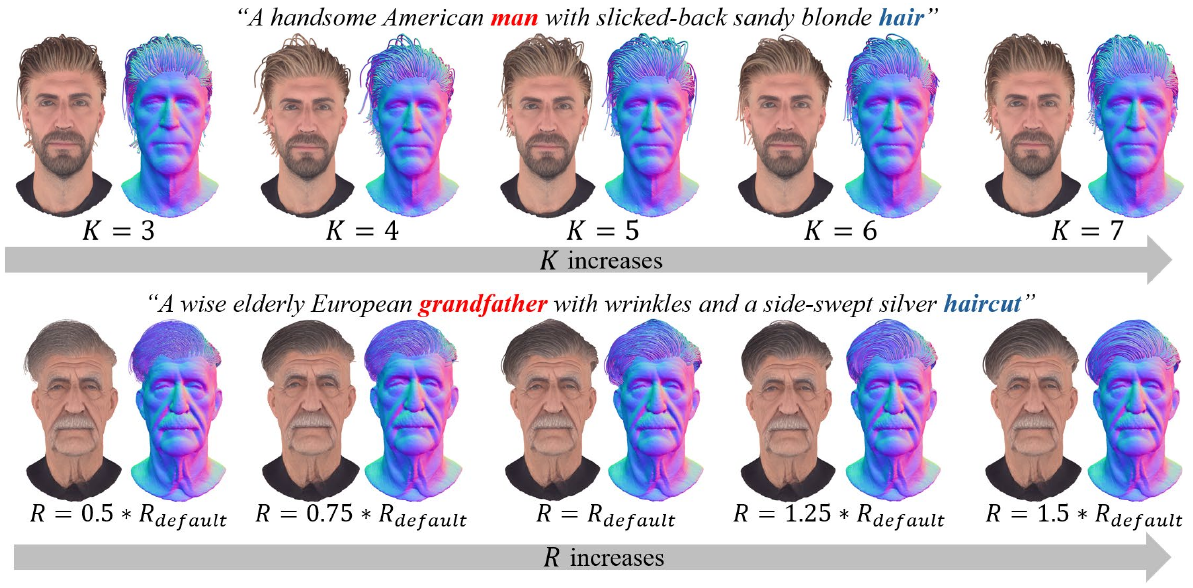} 
    \vspace{-0.3cm}
    \caption{Ablation study on $K$ and $R$.}
    \label{fig:K_R}
    \vspace{-0.3cm}
\end{figure}

\noindent
\textbf{Details of Differentiable Prismatization Algorithm.}
Given a hair strand $s$, our differentiable prismization algorithm converts it into a watertight prismatic mesh with $K$ lateral edges and radius $R$ through the following five steps:
\begin{enumerate}
    \item \textbf{Compute the Initial Normal Vector}: Determine a normal to the hair strand $s$ by taking the cross product of its orientation with a non-collinear reference point (typically the center of the head).
    \item \textbf{Generate $K$ Rotated Normals}: Rotate this normal around the axis defined by the strand's orientation $K$ times, each by $\frac{360^\circ}{K}$, to produce $K$ normals.
    \item \textbf{Translate to Form Lateral Edges}: Translate $s$ along each of these $K$ normals by $R$, generating $K$ lateral edges of the prism.
    \item \textbf{Construct Lateral Faces}: Connect the adjacent lateral edges'vertices to form the $K$ lateral faces of the prism.
    \item \textbf{Construct Top and Bottom Faces}: Connect the vertices at the ends of the lateral edges to form the top and bottom faces of the prism, completing the conversion from a hair strand to a watertight prismatic mesh.
\end{enumerate}
\Cref{fig:dp_pipeline} shows an example of converting a hair strand into a prismatic mesh.
Importantly, the proposed differentiable prismatization algorithm can be easily implemented on GPU, achieving flexible and fast prismatization of hair strands and paving a new way for hair modeling.

Specific to our experiment, each hair strand is converted into a watertight prismatic mesh with $K = 4$ lateral edges. The radius is defined as $R = \sqrt{\frac{A_{\text{scalp}}}{N_{s}\pi}}$, where $A_{\text{scalp}}$ represents the surface area of the scalp mesh. During the optimization of hair textures, the radius $R$ is further reduced to $\frac{\sqrt{\frac{A_{\text{scalp}}}{N_{s}\pi}}}{2}$ to achieve a more detailed appearance.
As illustrated in \cref{fig:dp}, our proposed differentiable prismatization algorithm offers more stable gradient backpropagation compared to the quad mesh used by NeuralHaircut~\cite{neural_haircut}. This mitigates abnormal normal problems caused by non-watertight meshes, ensuring reliable strand-based hair optimization.
Additionally, we present an ablation study on the effect of the number of hair strands in \cref{fig:strand_number}. 
It is observed that as the number of hair strands increases, the generated hairstyles become denser. Meanwhile, the overall shape and texture quality are maintained, demonstrating the robustness of differentiable prismatization.
As shown in \cref{fig:K_R}, we also conduct an ablation study on lateral edges $K$ and the radius $R$.
The results indicate that the number of lateral edges $K$ has a minimal impact on the final output, which validates the effectiveness and robustness of our differentiable prismatization algorithm.
    


\noindent
\textbf{Details of Geometry-Aware Losses.}
The geometry-aware loss functions $\mathcal{L}_{\text{bbox}}$, $\mathcal{L}_{\text{face}}$, and $\mathcal{L}_{\text{colli}}$ are formulated as:
\begin{align}
\mathcal{L}_{\text{bbox}} &= \sum_{p \in S} \max(0, s_{\text{bbox}}(p)),\\
\mathcal{L}_{\text{face}} &= \sum_{p \in S} \max(0, -s_{\text{face}}(p)),\\
\mathcal{L}_{\text{colli}} &= \sum_{p \in S} \max(0, -s_{\text{head}}(p)),
\end{align}
where $s_{\text{bbox}}$, $s_{\text{face}}$, and $s_{\text{head}}$ are the SDF of the bounding box, the space in front of the face, and the head, respectively. Here, $p$ represents the 3D points of hair strands $S$.

\begin{figure}[tp]
    \centering
    \includegraphics[width=0.43\textwidth]{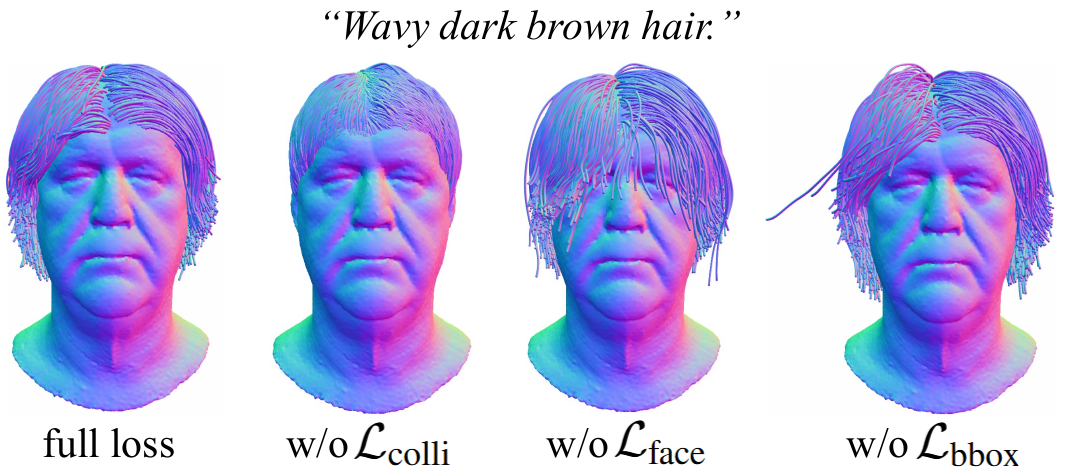} 
    \vspace{-0.3cm}
    \caption{Ablation study on geometry-aware losses.}
    \label{fig:geo_loss}
    \vspace{-0.3cm}
\end{figure}

The results of the ablation study on geometry-aware losses are displayed in \cref{fig:geo_loss}. Incorporating $\mathcal{L}_{\text{bbox}}$, $\mathcal{L}_{\text{face}}$, and $\mathcal{L}_{\text{colli}}$ effectively prevents the hair from extending beyond the bounding box, obscuring the face, and colliding with the head. 
This significantly enhances the geometric rationality and realism of the generated hairstyles.

\begin{figure}[tp]
    \centering
    \includegraphics[width=0.43\textwidth]{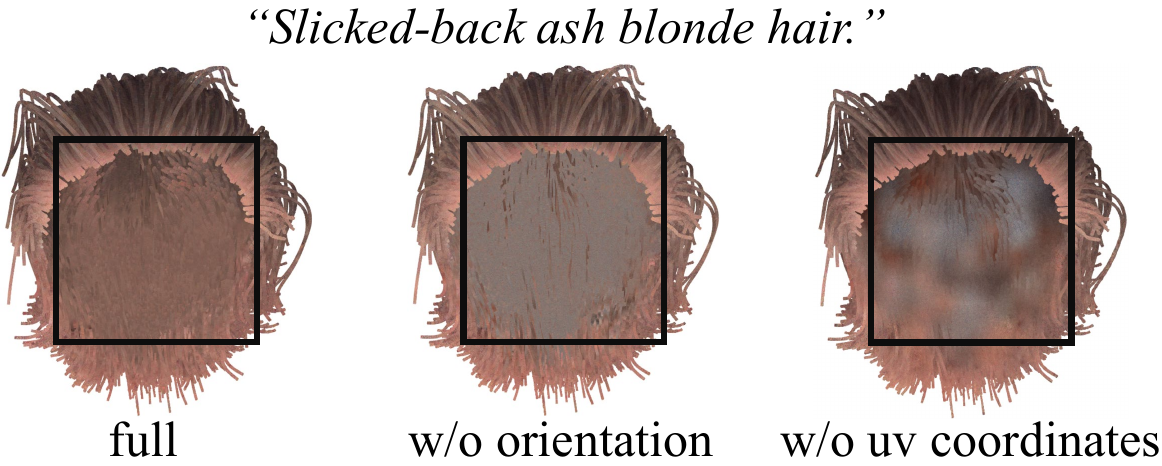} 
    \vspace{-0.3cm}
    \caption{Ablation study on the strand-aware texture field.}
    \label{fig:texture_field}
    \vspace{-0.3cm}
\end{figure}

\noindent
\textbf{Details of Strand-Aware Texture Field.}
Due to the complexity of hair textures, a basic texture field cannot fully capture the lifelike appearance of strands.
For a query point $p$, the basic texture field $\phi_\text{b}$ generates its color using the following formula:
\begin{equation}
    c = \phi_\text{b}(Euc(p)),
\end{equation} 
where $Euc(\cdot)$ represents the Euclidean coordinates of the query point.
To better model high-frequency color variations, we propose the strand-aware texture field $\phi_\text{s}$ which uses the following equation:
\begin{equation}
    c = \phi_\text{s}(UV(p), o),
\end{equation} 
where $UV(\cdot)$ denotes the scalp UV coordinates of the query point, and $o$ refers to its strand orientation.

Specifically, we introduce two improvements to the basic texture field: first, we replace Euclidean coordinates with scalp UV coordinates, which are more uniformly distributed.
Second, we incorporate strand orientations as additional input information to model orientation-dependent texture variations. 
These enhancements enhance the realism of the generated results by better capturing high-frequency appearance variations and ensuring consistent colors across different faces of a single prismatic mesh since the input features are strand-based.
As shown in \cref{fig:texture_field}, our proposed strand-aware texture field accurately models high-frequency appearance details by switching coordinate spaces and incorporating orientation information, resulting in more realistic strand textures.

\begin{figure}[tp]
    \centering
    \includegraphics[width=0.43\textwidth]{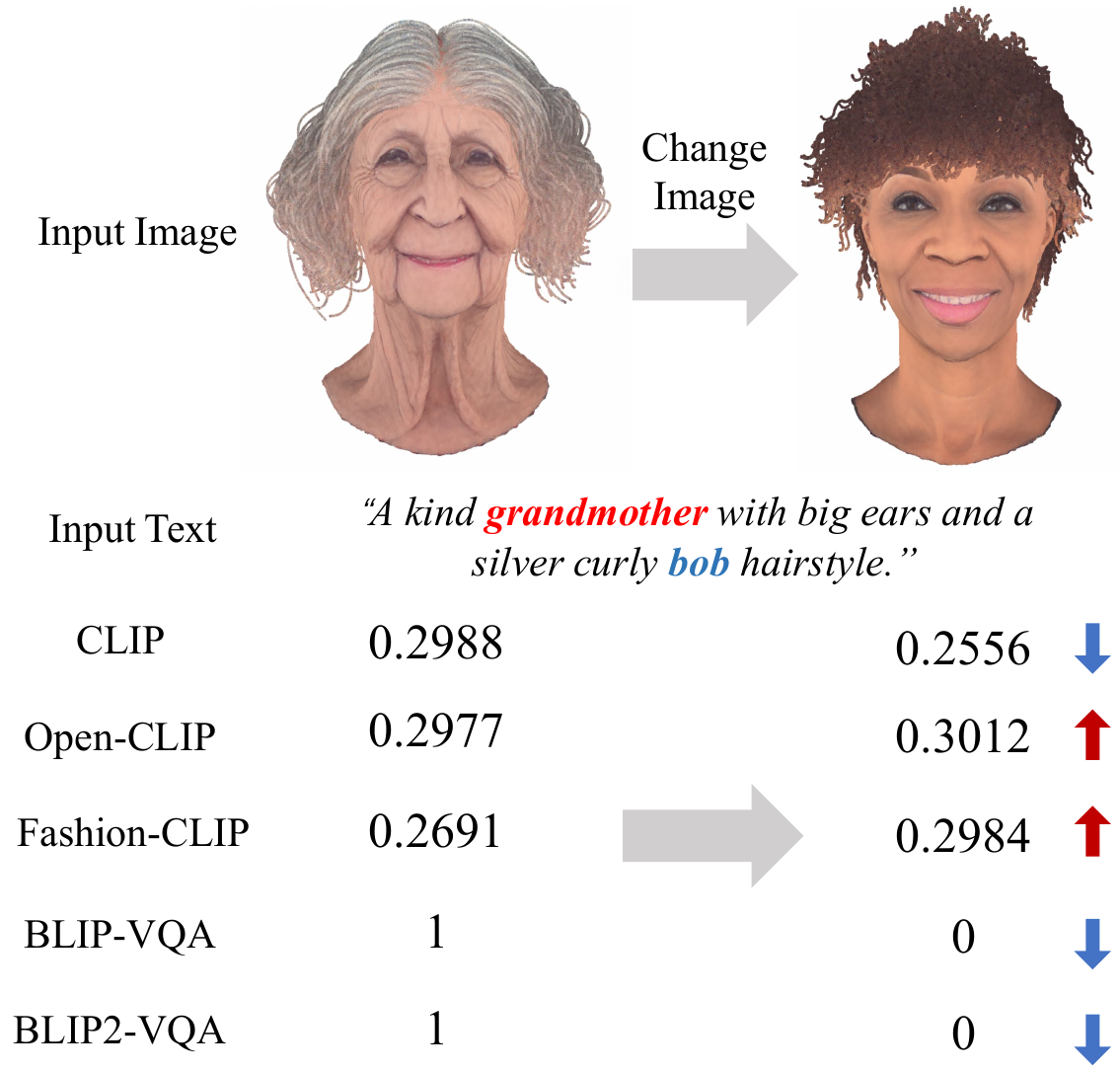} 
    \vspace{-0.3cm}
    \caption{Quantitative comparisons for metrics including CLIP, Open-CLIP, Fashion-CLIP, BLIP-VQA, and BLIP2-VQA.}
    \label{fig:clip}
    \vspace{-0.3cm}
\end{figure}

\begin{figure}[tp]
    \centering
    \includegraphics[width=0.43\textwidth]{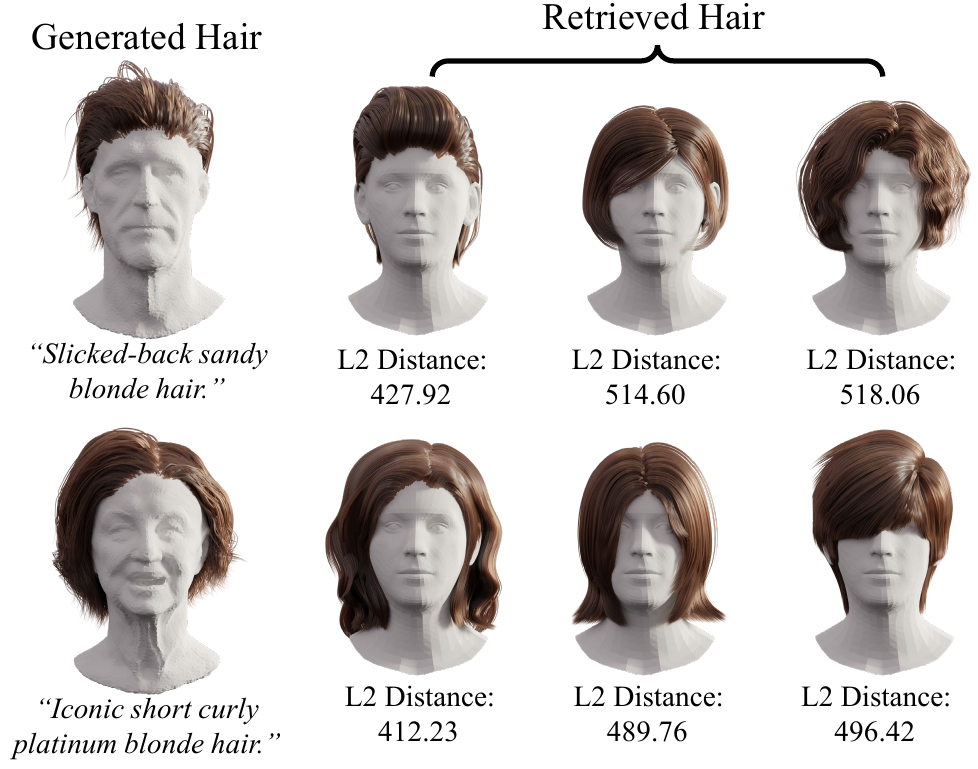} 
    \vspace{-0.3cm}
    \caption{The generated hair and the retrieved hair from USC-HairSalon~\cite{sing_view_rec_hair} by ranking the L2 distance of neural scale textures.}
    \label{fig:result_3}
    \vspace{-0.3cm}
\end{figure}

\subsection{More Evaluations}
For a more comprehensive comparison, we also provide evaluations based on FID and CLIP metrics.
We compute the FID between the images rendered by each method and those generated by Flux.
As shown in \cref{tab:quantitative_comparisons2}, our method still achieves good results.
However, given the inherent flaws of these metrics, we recommend that readers refer to the metrics in the main paper.

\begin{table}[tp]
    \centering
    \resizebox{0.5\linewidth}{!}{
        \begin{tabular}{ccccc}
            \toprule
            Method & FID $\downarrow$ & CLIP $\uparrow$ \\
            \midrule
            HeadArtist & 201.35 & 27.69 \\
            HeadStudio & 271.37 & 27.61 \\
            HumanNorm & 211.65 & 27.96 \\
            TECA & \underline{178.79} & \underline{30.60} \\
            StrandHead (Ours) & \textbf{176.95} & \textbf{30.95} \\
            \midrule
            MVDream & 215.06 & \textbf{28.11} \\
            GaussianDreamer & 249.96 & 25.05 \\
            LucidDreamer & 231.65 & 25.05 \\
            RichDreamer & \underline{213.54} & 27.13 \\
            TECA & 231.23 & 26.36 \\
            HAAR & 335.07 & 25.42 \\
            StrandHead (Ours) & \textbf{201.19} & \underline{27.84} \\
            \bottomrule
        \end{tabular}
    }
    \vspace{-0.3cm}
    \caption{Quantitative comparisons with the SOTA methods. The best and second-best results are highlighted in \textbf{bolded} and \underline{underlined}, respectively.} 
    \label{tab:quantitative_comparisons2}
    \vspace{-0.6cm}
\end{table}

\subsection{More Experiment Details and Results}
\textbf{Disadvantages of CLIP-Based Metrics.}
Recent studies~\cite{t2i-compbench,llmscore} have demonstrated that CLIP-based metrics are limited to assessing coarse text-image similarity and struggle to capture the fine-grained correspondence between 3D content and input prompts accurately.
To address this limitation, we follow Progressive3D~\cite{progressive3d} and utilize fine-grained text-to-image evaluation metrics, such as BLIP-VQA~\cite{blip,lavis} and BLIP2-VQA~\cite{blip2,lavis}, to assess the generative capabilities of different methods.

As depicted in \cref{fig:clip}, when input images are replaced while keeping the input text unchanged, interesting observations can be made: Open-CLIP and Fashion-CLIP scores increase under these conditions. However, BLIP-VQA and BLIP2-VQA scores drop significantly. This highlights the limitations of CLIP-based metrics in evaluating fine-grained correspondences. In contrast, BLIP-VQA and BLIP2-VQA demonstrate superior performance in capturing these intricate relationships.

\noindent
\textbf{Clarification on Retrieval-Like Results.}
Since there is no GT available for text-driven hair generation, we display the generated hair alongside its top-3 nearest haircuts from the USC-HairSalon dataset~\cite{sing_view_rec_hair}textures, as shown in \cref{fig:result_3}. The similarity is ranked based on the L2 distance of neural scale textures. As illustrated, the generated hair is significantly different from the retrieved hair samples. This distinction clearly demonstrates that our method generates unique hairstyles rather than simply retrieving them from the dataset.

\begin{figure*}[tp]
    \centering
    \includegraphics[width=1\textwidth]{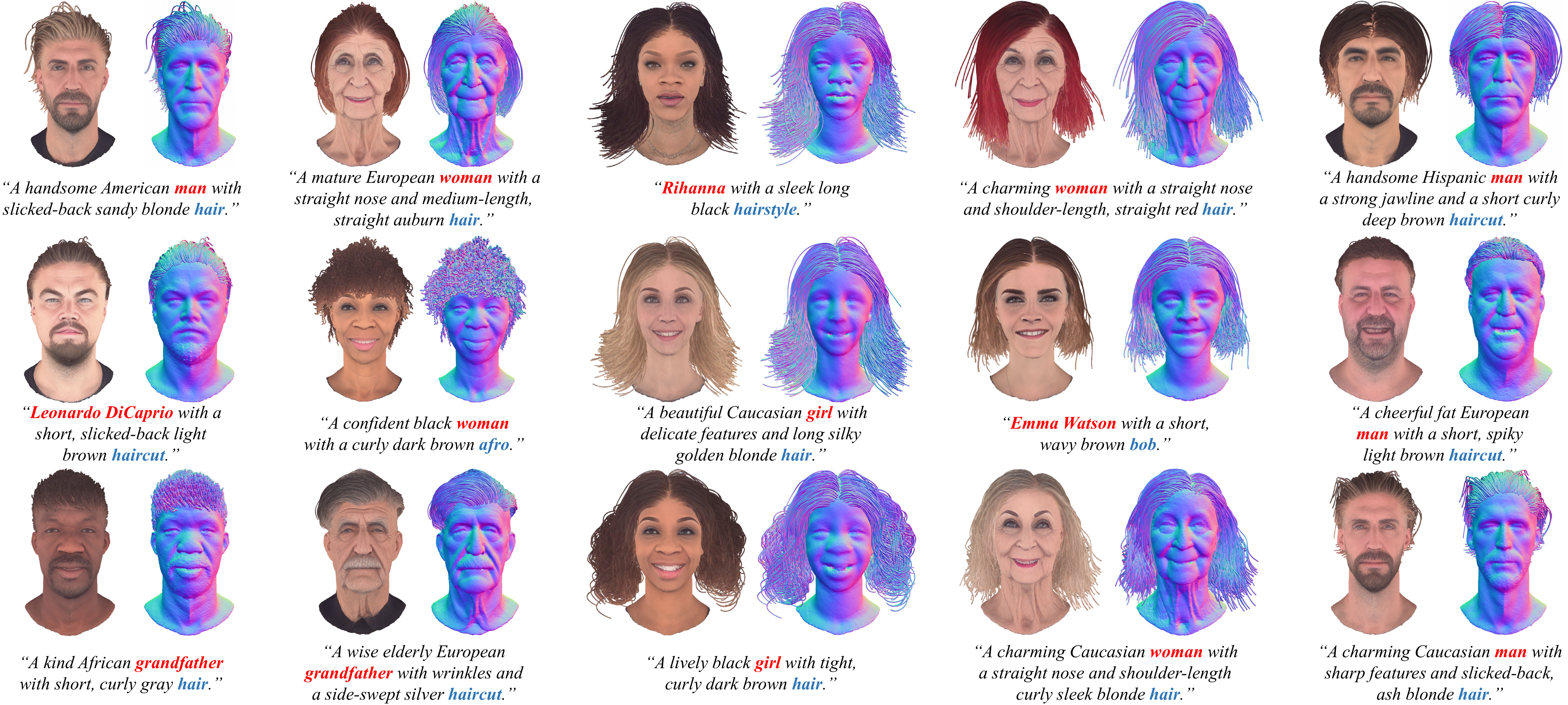} 
    \vspace{-0.3cm}
    \caption{Examples of 3D head avatars with strand-based attributes.}
    \label{fig:result_1}
    \vspace{-0.3cm}
\end{figure*}

\begin{figure*}[tp]
    \centering
    \includegraphics[width=1\textwidth]{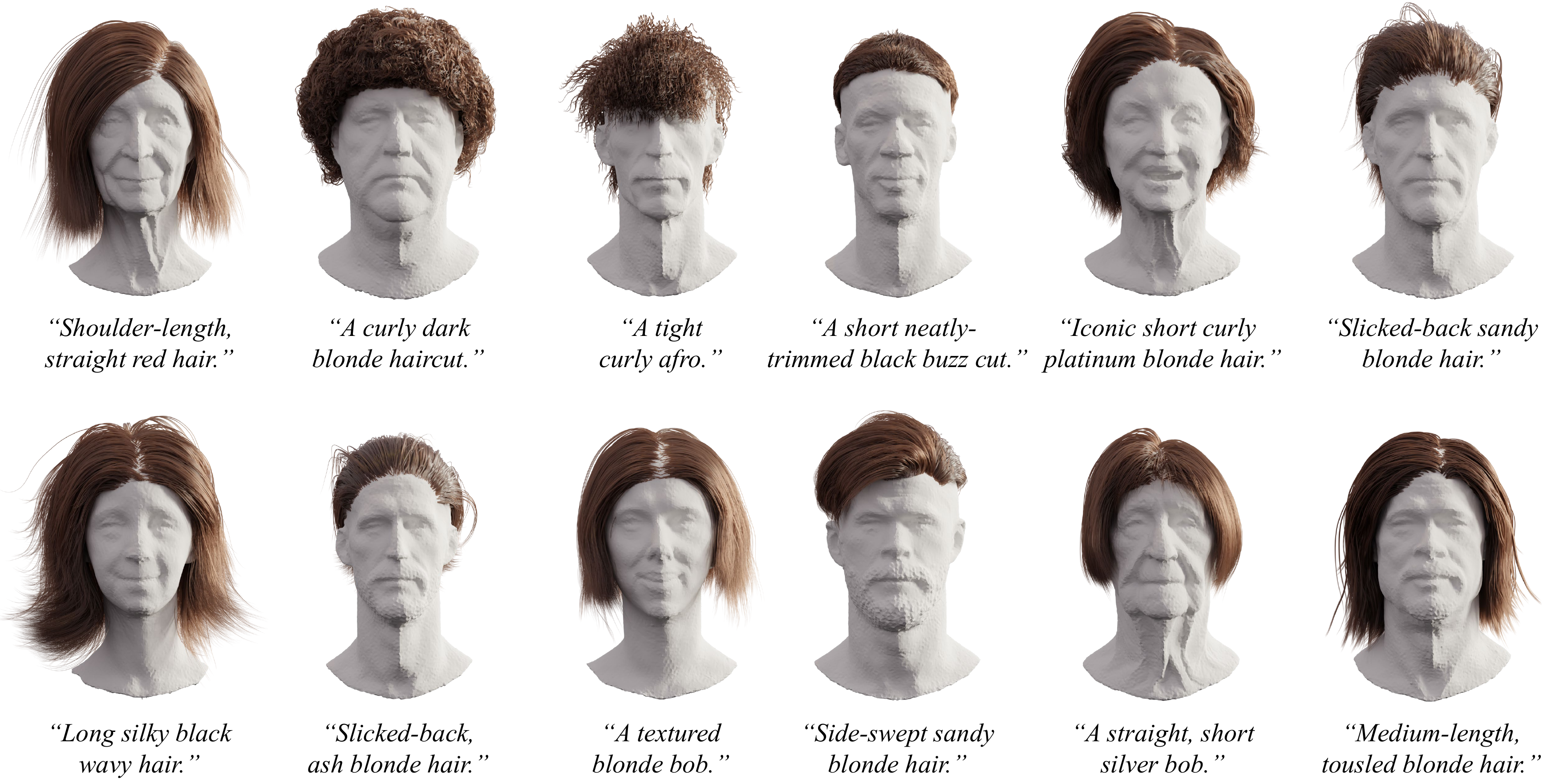} 
    \vspace{-0.3cm}
    \caption{The physics-based hair strand rendering result using Blender~\cite{blender}.}
    \label{fig:result_2}
    \vspace{-0.3cm}
\end{figure*}

\noindent
\textbf{More Experiment Results.}
We present additional 3D hair-disentangled head avatars in \cref{fig:result_1} and 3D strand-based hair generated by StrandHead in \cref{fig:result_2}. 

\begin{figure*}[tp]
    \centering
    \includegraphics[width=0.95\textwidth]{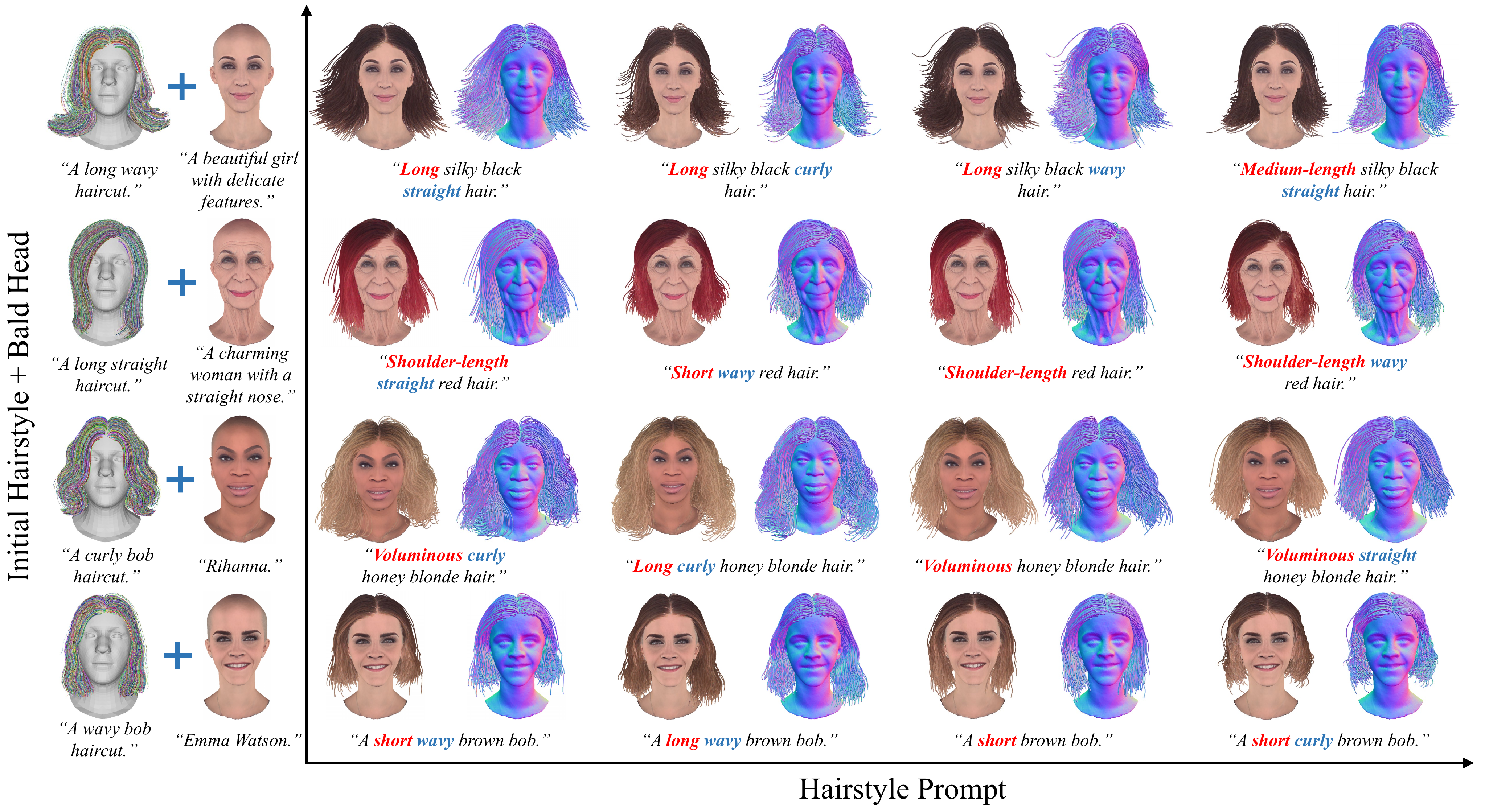} 
    \vspace{-0.3cm}
    \caption{The optimization results under different hair prompts. Starting from the same initial hairstyle, StrandHead demonstrates its capability to generate diverse hairstyles by adapting to varying text conditions.}
    \label{fig:diff_hair}
    \vspace{-0.3cm}
\end{figure*}

\noindent
\textbf{Effect of Human-Specific 2D Generative Priors.}
We demonstrate the importance of human-specific 2D generative priors from two aspects:

\textbf{(1)} \cref{fig:diff_hair} displays optimized hair under varying text conditions, while keeping the initial hairstyle and bald head constant. As shown, our method, leveraging human-specific 2D generative priors, accurately captures subtle changes in textual descriptions (e.g., variations in haircut length and curliness). The generated 3D strand-based hair not only exhibits a realistic and well-structured shape but also aligns closely with the given text conditions.

\begin{figure*}[tp]
    \centering
    \includegraphics[width=0.95\textwidth]{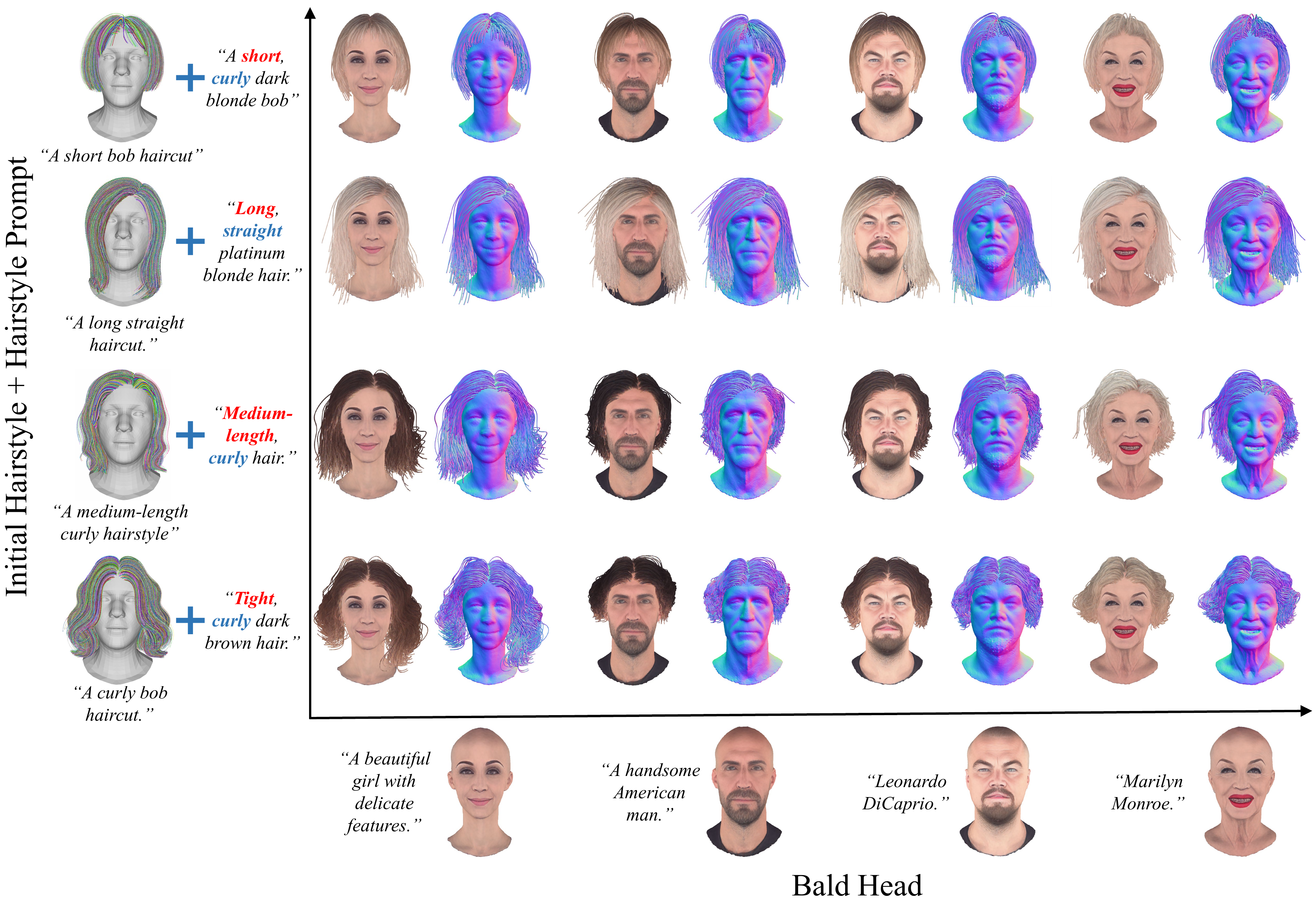} 
    \vspace{-0.3cm}
    \caption{The optimization results under different bald heads. Under consistent text conditions, the 3D hair generated by StrandHead exhibits specific geometry and texture variations that seamlessly adapt to the unique features of different bald heads.}
    \label{fig:diff_head}
    \vspace{-0.3cm}
\end{figure*}

\textbf{(2)} \cref{fig:diff_head} illustrates generated hair under varying bald head conditions while maintaining a fixed initial hairstyle and hair prompt. As observed, thanks to the powerful 2D diffusion models pre-trained on human data, the generated 3D hair strands exhibit geometry and texture variations that adapt to specific bald heads.

In summary, our robust optimization strategy ensures that the generated hair is not only highly consistent with the text prompts but also integrates harmoniously with the human head.

\subsection{Prompt List}
The following are textual prompts for quantitative experiments:
\begin{itemize}
    \item A beautiful girl with delicate features and long, silky black wavy hair.
    \item A cheerful fat European man with a short, spiky light brown haircut.
    \item A confident black woman with a curly dark brown afro.
    \item A handsome American man with slicked-back sandy blonde hair.
    \item A kind grandmother with big ears and a silver curly bob hairstyle.
    \item A lively European boy with tousled light brown hair.
    \item A lively black girl with tight, curly dark brown hair.
    \item A mature African man with deep-set eyes and a short, curly black afro.
    \item A mature European woman with a straight nose and medium-length, straight auburn hair.
    \item A middle-aged Hispanic woman with a strong jawline and medium-length, wavy magenta hair.
    \item A muscular European man with a wide forehead and slicked-back black hair.
    \item A serious white man with a sharp nose and a slicked-back, jet-black hairstyle.
    \item A sexy woman with full lips and long, wavy chestnut brown hair.
    \item A strong American man with a gray beard and a curly silver haircut.
    \item A strong man with a broad nose and a short, spiky dark brown haircut.
    \item A strong man with a strong jaw and a medium-length, wavy black hairstyle.
    \item A stylish African woman with a sleek, shoulder-length black bob haircut.
    \item A thin man with a sharp jawline and a spiky light brown mohawk hairstyle.
    \item A wise elderly European grandfather with wrinkles and a side-swept silver haircut.
    \item A wise, elderly grandfather with a classic short curly white haircut.
    \item An elderly African grandmother with wrinkles and a curly gray afro.
    \item An elderly Caucasian grandmother with thin lips and a straight, short silver bob.
    \item Angelina Jolie with long, wavy chestnut brown hair.
    \item Beyoncé with voluminous, curly honey blonde hair.
    \item Emma Watson with a short, wavy brown bob.
    \item Lionel Messi with a blonde mohawk hairstyle.
    \item Michael Jordan with a short, neatly-trimmed black buzz cut.
    \item Morgan Freeman with a curly brown afro.
    \item Rihanna with a sleek, long black hairstyle.
    \item Taylor Swift with long, straight platinum blonde hair.
\end{itemize}

\subsection{Ethics Statement}
StrandHead provides an efficient solution for creating realistic 3D head avatars with strand-based hair using 2D/3D human-centric priors, enabling a wide range of applications. However, like many AI generative technologies, it carries the risk of misuse, such as the creation of misleading avatars.
To mitigate these concerns, future research in generative AI should emphasize ethical considerations, develop effective safeguards, and promote responsible practices. By addressing these challenges, developers can reduce potential harm while maximizing the positive impact of these technologies.

\end{document}